\documentclass[12pt]{article}

\usepackage{arxiv}
\usepackage{amsmath}
\usepackage{amsthm}
\usepackage[utf8]{inputenc} 
\usepackage[T1]{fontenc}    
\usepackage{hyperref}       
\usepackage{url}            
\usepackage{booktabs}       
\usepackage{amsfonts}       
\usepackage{nicefrac}       
\usepackage{microtype}      
\usepackage{cleveref}       
\usepackage{lipsum}         
\usepackage{graphicx}
\usepackage{natbib}
\usepackage{doi}
\usepackage{comment}

\title{The Uniqueness of LLaMA3-70B Series with Per-Channel Quantization}

\newif\ifuniqueAffiliation
\uniqueAffiliationtrue

\ifuniqueAffiliation 
\author{ Minghai Qin \\
	Western Digital Research\\
	\texttt{minghai.qin@wdc.com} 
}
\else
\usepackage{authblk}

\newbox{\orcid}\sbox{\orcid}{\includegraphics[scale=0.06]{orcid.pdf}} 
\author[1]{%
	\href{https://orcid.org/0000-0000-0000-0000}{\usebox{\orcid}\hspace{1mm}David S.~Hippocampus\thanks{\texttt{hippo@cs.cranberry-lemon.edu}}}%
}
\author[1,2]{%
	\href{https://orcid.org/0000-0000-0000-0000}{\usebox{\orcid}\hspace{1mm}Elias D.~Striatum\thanks{\texttt{stariate@ee.mount-sheikh.edu}}}%
}
\affil[1]{Department of Computer Science, Cranberry-Lemon University, Pittsburgh, PA 15213}
\affil[2]{Department of Electrical Engineering, Mount-Sheikh University, Santa Narimana, Levand}
\fi

\hypersetup{
pdftitle={A template for the arxiv style},
pdfsubject={q-bio.NC, q-bio.QM},
pdfauthor={David S.~Hippocampus, Elias D.~Striatum},
pdfkeywords={First keyword, Second keyword, More},
}

\usepackage[normalem]{ulem}

\usepackage{soul}
\newif\ifmodify

\ifmodify

\newcommand{\del}[1]{\textcolor{gray}{\sout{#1}}}
\newcommand{\chatgpt}[1]{\textcolor{blue}{#1}}
\newcommand{\claude}[1]{\textcolor{brown}{#1}}

\else

\newcommand{\del}[1]{}
\newcommand{\claude}[1]{#1}
\newcommand{\chatgpt}[1]{}

\fi

\begin{document}
\maketitle

\begin{abstract}
We have observed a distinctive quantization-related behavior in the LLaMA3/3.1-70B models that is absent in both the LLaMA2-70B and LLaMA3/3.1/3.2-1B/3B/8B/405B models. 
Quantization is a crucial technique for deploying large language models (LLMs) efficiently, as it reduces memory consumption, decreases memory transactions, and potentially accelerates inference using lower-precision compute cores. Among various bit widths and representations for weights and activations, the 8-bit integer weight and 8-bit integer activation (W8A8) configuration is particularly popular due to its widespread hardware support. However, the impact of W8A8 post-training quantization on model accuracy, especially on the recently released LLaMA3/3.1 model series, remains contentious. While several studies have suggested calibrating either weights or activations to mitigate accuracy degradation, a comprehensive solution has yet to be identified.
In this paper, we explore three key questions: What makes the LLaMA3-70B model series uniquely vulnerable to quantization? Why is this the case? And how can the issue be addressed? We empirically investigate multiple LLMs featured on an open LLM leaderboard, discovering that the LLaMA3-70B model series have a \textbf{unique} accuracy degradation behavior with W8A8 per-channel post-training quantization. In contrast, other model series such as LLaMA2, LLaMA3/3.1-8B, LLaMA3.2-1B/3B, Qwen, Mixtral, Mistral, Phi-3, and Falcon demonstrate robust performance with W8A8, sometimes surpassing their FP16 counterparts.
Contrary to previous assertions attributing degradation to the large dynamic range of activations, our findings indicate that the weight distribution of the LLaMA3-70B is the primary factor behind the vulnerability. By meticulously analyzing the distinct characteristics of weight distributions across Transformer blocks, we propose two solutions that make different tradeoffs in hardware/software overhead. 
First, we propose a mixed strategy where less than 3\% of the layers employ finer per-group W8A8 quantization granularity, while the remaining 97\% retain the per-channel configuration. Second, we introduce a bi-smoothing strategy that balances quantization errors between weights and activations while maintaining per-channel quantization throughout.
Experimental results demonstrate that both strategies effectively preserve the accuracy of the entire LLaMA3-70B model series under W8A8 quantization, achieving performance on par with their FP16 counterparts.

\end{abstract}

\keywords{LLaMA models, quantization}

\section{Introduction}

The rising popularity of large language models (LLMs)~\cite{vaswani2023attentionneed,brown2020languagemodelsfewshotlearners,devlin2019bertpretrainingdeepbidirectional,liu2019robertarobustlyoptimizedbert,touvron2023llamaopenefficientfoundation} in diverse practical applications underscores their revolutionary impact on natural language processing, data analysis, and artificial intelligence. However, their immense size imposes a substantial burden on GPU memory~\cite{kaplan2020scalinglawsneurallanguage}, complicating deployment in resource-constrained environments. Quantization is a pivotal technique to mitigate this issue, reducing the memory footprint of LLMs while maintaining accuracy~\cite{gholami2021surveyquantizationmethodsefficient,Zafrir_2019,wang2023bitnetscaling1bittransformers}. By converting high-precision weights and activations to lower-precision formats, quantization facilitates more efficient model storage and accelerates inference. As the demand for LLMs grows, the advancement of quantization methods will be crucial for their scalable and cost-effective deployment in real-world applications.

Quantization uses lower precision to represent weights and activations. In this study, we focus on post-training integer-bit quantization (PTQ). The 8-bit weights and 8-bit activations (W8A8) configuration is often regarded as a good balance between efficiency and accuracy. Compared to the FP16 counterpart, W8 can save reduce the memory consumption and transaction by 50\%. Although 4-bit weights and 16-bit activations (W4A16) require even less memory, they often lack broad hardware support for the INT4-FP16 matrix-multiplication compute cores. In contrast, existing INT8-INT8 compute cores offer superior acceleration for the inference of W8A8 models. On the other hand, while the W4A4 configuration may reduce memory usage and latency through INT4-INT4 compute cores, it typically suffers from significant accuracy degradation due to the aggressive quantization. 

The granularity of a quantization group in large language models (LLMs) is another critical factor influencing the accuracy-acceleration trade-off. Within a quantization group, all values share the same scaling factor and zero point. When the group size matches the input dimension of a layer, this is termed ``per-channel'' quantization, which maximizes potential speedup. Conversely, ``per-group'' quantization occurs when an input channel is divided into multiple groups. This requires the intermediate partial sums of matrix-matrix multiplication to be transferred out of the compute cores, significantly reducing system throughput. Therefore, maintaining accuracy with per-channel quantization is essential for practical applications.

\begin{figure}[tbp]
    \centering

    \includegraphics[width=1\linewidth]{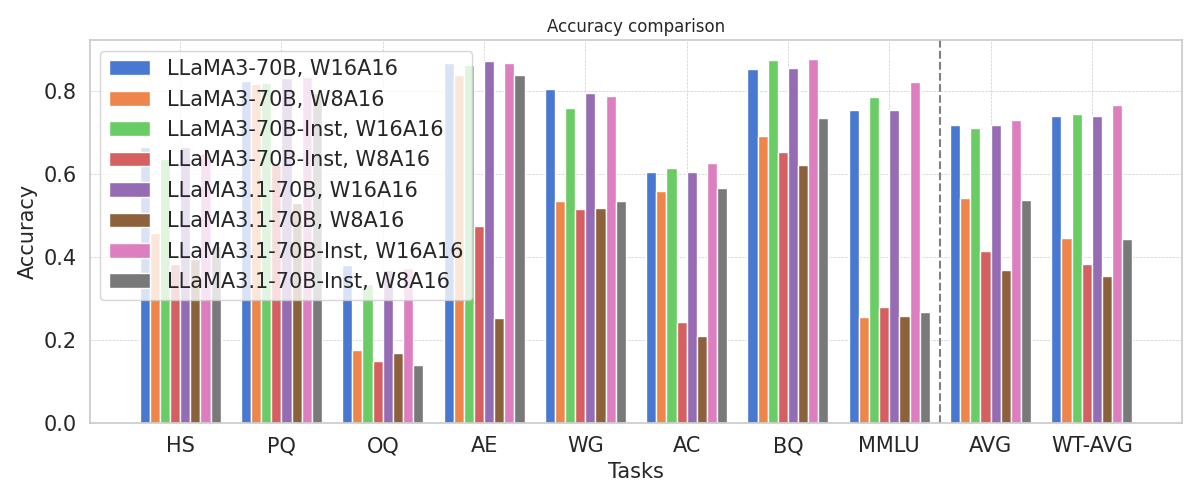}
    \caption{Accuracy  of LLaMA3-70B model series with W16A16, and W8A16 per-channel quantization}
    
    \label{fig:llama3-70b-w8a8}
\end{figure}

In this paper, we investigate the W8A8 quantization for the most popular architectures and models featured on the LLM open leaderboard~\cite{open-llm-leaderboard}. Our study reveals a noteworthy phenomenon: the LLaMA3-70B model series~\cite{dubey2024llama3herdmodels} exhibits a pronounced vulnerability to W8A8 per-channel quantization, in stark contrast to other models, which demonstrate significantly greater robustness, typically experiencing less than 1\% accuracy degradation compared to their FP16 counterparts under the same quantization scheme. 
Our observation on the ``LLaMA3-70B series'' also include the LLaMA3.1-70B and various fine-tuned versions, and other robust models include LLaMA2 series, LLaMA3-8B, Mistral-123B, Mixtral-8x7B, Qwen2-72B, Phi3-13B, and Falcon-40B~\cite{jiang2023mistral7b,jiang2024mixtralexperts,yang2024qwen2technicalreport,abdin2024phi3technicalreporthighly,almazrouei2023falconseriesopenlanguage}.  Figure~\ref{fig:llama3-70b-w8a8} provides a comparative analysis of FP16 (labeled as W16A16) and W8A16 quantization for the officially released LLaMA3/3.1-70B and LLaMA3/3.1-70B-Instruct by Meta. This comparison reveals that significant accuracy degradation already occurs with W8A16, indicating that the issue primarily stems from weight quantization rather than activation quantization. Details will be presented in the experiment section. Released in April 2024, the LLaMA3-70B has not been extensively examined in recent research on W8A8 per-channel quantization. Given that this series is currently the most widely used base model on the open leaderboard, further exploration and understanding of its unique characteristics are imperative.

\begin{figure}[tbp]
    \centering
    \includegraphics[width=0.7\linewidth]{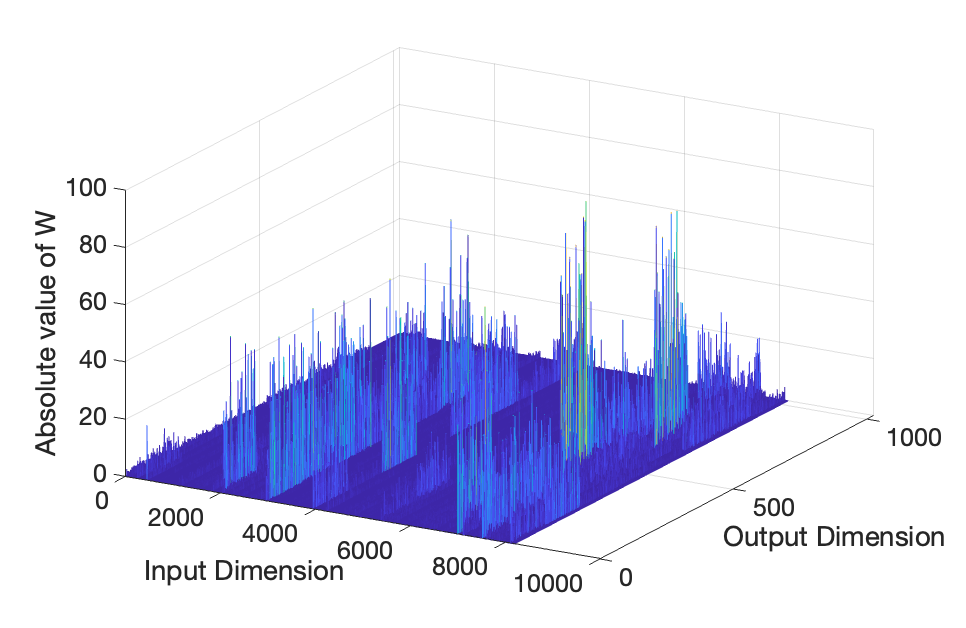}
    \caption{abs\_weights of the ``V'' matrix in the first Transformer block of LLaMA3-70B}
    \label{fig:wd_llama3-70b}
\end{figure}

To elucidate the unique vulnerability of the LLaMA3-70B series to W8A8 per-channel quantization, we conducted a comparative analysis of its weight distributions against those of other robust models. Our investigation revealed a significant discrepancy, particularly pronounced in the initial layers. These layers exhibit a distinctive characteristic wherein magnitudes of certain weights exceed others by several orders of magnitude. Figure~\ref{fig:wd_llama3-70b} shows an example of the ``V'' matrix in the first Transformer block of the LLaMA3-70B model. Weights with large magnitudes are concentrated on specific input dimensions, with the largest values exceeding 90.
These outlier weights substantially expand the quantization range, resulting in larger quantization intervals and consequently diminished precision for smaller weight values. Notably, these weight outliers are only prominent in the ``Q'', ``K'', ``V'', ``Up'', and ``Gate'' weight matrices of the initial transformer blocks, while the "O" and "Down" weight matrices do not exhibit this behavior. 
We propose two methods to address this issue. First, to mitigate these quantization errors, we propose implementing per-group quantization specifically for these affected layers. Although these layers constitute merely 2-3\% of the total layers in LLaMA3-70B, this targeted approach yields a dramatic enhancement in accuracy, elevating model performance to a level approaching that of FP16 models. Second, we demonstrate a bi-directional smoothing method of the weight and activation matrices that effectively reduces the dynamic range of both, significantly minimizing quantization errors. This technique preserves per-channel quantization while enhancing the model's resilience to W8A8 quantization.

The contribution of this paper is as follows.

\begin{itemize}
\item We show that the LLaMA3-70B model series has a unique characteristic that makes it the only model series on the open LLM leaderboard vulnerable to W8A8 per-channel post-training quantization (PTQ). In contrast, all other models can be safely quantized to per-channel-W8A8 with negligible accuracy loss. We believe this observation holds significant implications since LLaMA3-70B serves as one of the most popular base models for numerous models fine-tuned for domain-specific tasks.
\item We identify that LLaMA3-70B's sensitivity to per-channel PTQ stems from its significantly different weight distribution, particularly the presence of substantial weight outliers.
\item To address this, we first propose a mixed per-channel/per-group quantization strategy. This approach applies per-group quantization to less than 3\% of the layers, specifically those with significant weight outliers, while maintaining per-channel quantization for the remaining 97\% layers. This mixed strategy effectively restores the accuracy of the LLaMA3-70B model series to levels comparable to its FP16 counterparts.
\item The second method introduces bi-directional smoothing to mitigate outliers in both weights and input activations. This approach significantly reduces the dynamic range required for quantization, enabling the LLaMA3-70B model series to retain accuracy comparable to their FP16 counterparts.
\end{itemize}

\section{Quantization Basics}
In this paper, we use symmetric integer-bit quantization due to its high efficiency and simpler hardware support. In symmetric $n$-bit integer quantization, we map floating-point (FP) values to integers in $[-2^{n-1}, 2^{n-1}-1]$. This process involves scaling the FP values based on a scale-factor and an offset. This offset is sometimes called ``zero-point'' and is set to $0$ for symmetric quantization. 

Let \( x \) be a FP value, and \( x_q \) its quantized integer representation. The scale-factor \( s \) is determined by the range of the FP values and the desired range of the integers. The symmetric quantization formula is given by:

\[
x_q = \text{round}\left(\frac{x}{s}\right)
\]
where \( \text{round}(\cdot) \) rounds the result to the nearest integer.

The scale-factor \( s \) is typically calculated based on the maximum absolute value of the FP values, \( \text{max\_abs} \), and the maximum representable integer value, \( 2^{n-1}-1 \). For symmetric quantization, the scale-factor is defined as:

\[
s = \frac{\text{max\_abs}}{2^{n-1}-1 }
\]

Thus, the quantization and dequantization processes can be expressed as:

\[
x_q = \text{round}\left(\frac{x}{s}\right)
\quad \text{and} \quad
x \approx x_q \cdot s
\]

The scale-factor $s$ plays a crucial role in the quantization process. The number of floating-point values used to determine \text{max\_abs} is typically referred to as the group size. In LLMs, consider a weight matrix $W$ of dimension $M\times N$  and an input activation matrix $A$ of dimension $L\times N$. The resultant output of $A\cdot W^T$  has dimension $L\times M$. Here $L$, $N$, and $M$ represent the sequence length, hidden dimension, and output dimension in a Transformer layer, respectively.
For per-channel quantization of $W$, the group size is $N$, implying that each row of $N$ values in $W$ shares a common scale-factor. Consequently, all scale-factors form a vector of length $M$. Similarly, per-channel quantization of $A$ also has a group size of $N$, with each row of $N$ values in $A$ sharing a scale-factor, resulting in a scale-factor vector of length $L$.
This configuration enables the computation of floating-point matrix multiplication using integer-bit compute cores (e.g., INT8 tensor cores), followed by dequantization. The dequantization process involves an element-wise multiplication with the scale-factor matrix of dimension $L\times M$, derived as the outer product of the scale-factor vectors of $W$ and $A$. Notably, this dequantization can often be efficiently fused into the subsequent operation after $A\cdot W^T$ in LLMs, which is typically an element-wise operation such as rotary embedding, SiLU activation application, or normalization.
In contrast, per-group quantization of $W$ and $A$ employs a finer granularity, where each row of $W$ (and of $A$) is partitioned into groups, each with its own scale-factor. Consequently, the output $A\cdot W^T$ cannot be computed in a single cycle within the INT tensor core, as the scale-factors differ during accumulation in the multiply-accumulation (MAC) unit. This leads to a degradation in the compute efficiency of the quantized model.
Given these considerations, per-channel quantization is generally preferred unless the quantization-induced error degrades the model's accuracy to an unacceptable level.

\section{The Uniqueness of LLaMA3-70B with per-channel W8A8}
\label{sec:experiment-llama3-70b}

\subsection{Test Dataset}
\del{We tested LLM models on 8 reasoning tasks, namely, Hellaswag (HS), PIQA (PQ), OpenbookQA (OQ), ARC-easy (AE), Winogrande (WG), ARC-Challenge (AE), BoolQ (BQ), and MMLU~. We observe that many studies report the average accuracy over reasoning task by simply averaging the accuracy of each task, which might not be rigorous since the number of questions in each task has large variance. For example, HS and MMLU both have over 10K questions while OQ has only 500 questions. In order to take this into account, we also report the ``weighted average accuracy'' (denoted by WT-AVG in figures), which is the total number of correct answers divided by the total number of questions. In this way, the accuracy of dataset with a large number of questions is weighted more than smaller dataset.}

\claude{We evaluated LLMs across eight reasoning tasks: Hellaswag (HS), PIQA (PQ), OpenbookQA (OQ), ARC-Easy (AE), Winogrande (WG), ARC-Challenge (AC), BoolQ (BQ), and MMLU\cite{zellers2019hellaswagmachinereallyfinish,bisk2019piqareasoningphysicalcommonsense,OpenBookQA2018,allenai:arc,sakaguchi2019winograndeadversarialwinogradschema,clark2019boolqexploringsurprisingdifficulty,hendryckstest2021,hendrycks2021ethics}. We report the ``accuracy'' for all tasks. Some tasks have ``normalized accuracy'' and the conclusions derived from ``accuracy''  remain consistent when applied to ``normalized accuracy''. We realize that many studies report the average accuracy over these reasoning tasks by simply computing the arithmetic mean of individual task accuracies (denoted by ``AVG'' in our figures). However, this approach may lack rigor due to the significant variance in the number of questions across tasks. For instance, HS and MMLU each contain over 10,000 questions, while OQ comprises only 500.}

\claude{To address this disparity, we use an additional metric: the ``weighted average accuracy.'' This metric is calculated by dividing the total number of correct answers across all tasks by the total number of questions. Consequently, this method assigns greater weight to datasets with a larger number of questions, providing a more balanced representation of model performance across tasks of varying sizes. It is denoted by ``WT-AVG'' in our figures.}

\subsection{Tested Models}
We  test the  accuracy of the following models: LLaMA3-70B, LLaMA3-70B-Instruct, LLaMA3.1-70B, Llama3-70B-Synthia, calme-2.2-llama3-70b, LLaMA3-8B, LLaMA2-70B, LLaMA2-70B-chat, Qwen2-72B, Mixtral-8x7B, Phi3-14B-Instruct, Mistral-L-Instruct-123B, Falcon-40B.  HuggingFace links to them will be provided in the appendix. We also analyze the weight distribution of LLaMA3.1-405B without evaluating its accuracy with FP16 due to GPU memory limitations.

\subsection{Test Environments}
All tests are done within 8xA100 (80GB) GPUs with PyTorch framework using lm-evaluation-harness~\cite{eval-harness}. 

\subsection{LLaMA3-70B's sensitivity to W8A8 }

\begin{figure}[tbp]
    \centering
    \includegraphics[width=1\linewidth]{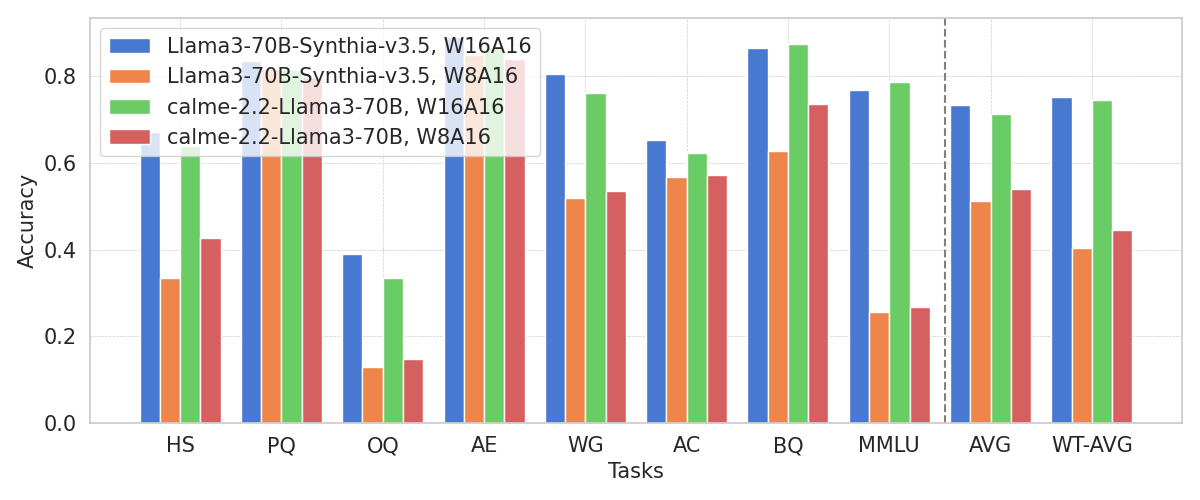}
    \caption{W8A8 per-channel quantization accuracy of models using  LLaMA3-70B as the base model }
    \label{fig:llama3-70b-finetuned-w8a16}
\end{figure}

\claude{Figure~\ref{fig:llama3-70b-w8a8} and Figure~\ref{fig:llama3-70b-finetuned-w8a16} present a comparative analysis of accuracy for several models in the LLaMA3-70B series subjected to 8-bit per-channel post-training quantization. This analysis encompasses the recently released LLaMA3.1-70B and two additional models fine-tuned from LLaMA3-70B. }
\claude{
These figures yield several notable observations:
\begin{itemize}
    \item The accuracy of models undergoes significant degradation with W8 quantization, even when activations are maintained at FP16 precision. This indicates that the observed accuracy deterioration is not attributable to quantization errors in activations, but rather originates from the 8-bit weight quantization process.
    \item Models fine-tuned from LLaMA3-70B exhibit the same vulnerability to W8 quantization as their base model. This persistence of vulnerability can be attributed to the fact that fine-tuning typically induces only minor alterations to weight values. In the subsequent section, we will demonstrate that this vulnerability is intrinsically linked to the underlying weight distributions of the model.
\end{itemize}
}

\begin{figure}[hthp]
    \centering
    \includegraphics[width=1\linewidth]{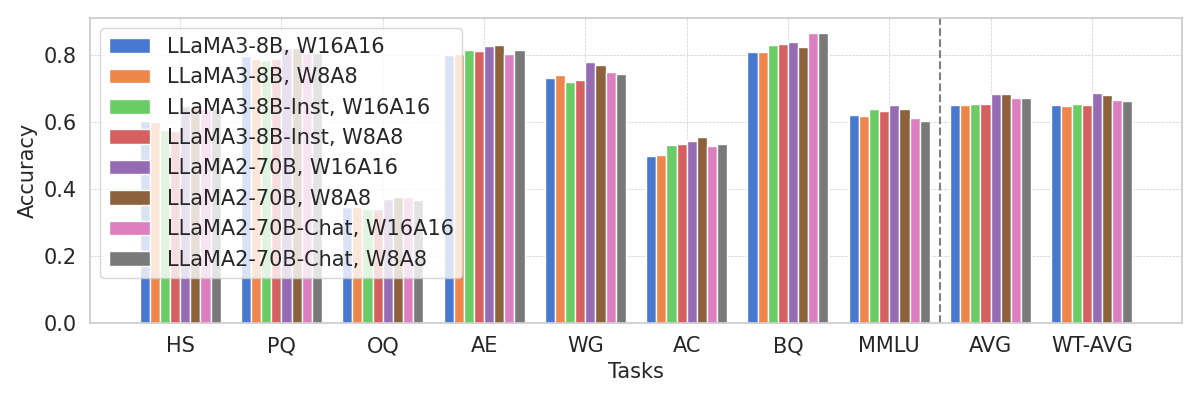}
    \caption{Accuracy of other LLaMA models  with W8A8 per-channel quantization}
    \label{fig:llama_else}
\end{figure}
  \begin{figure}[hthp]
    \centering
    \includegraphics[width=1\linewidth]{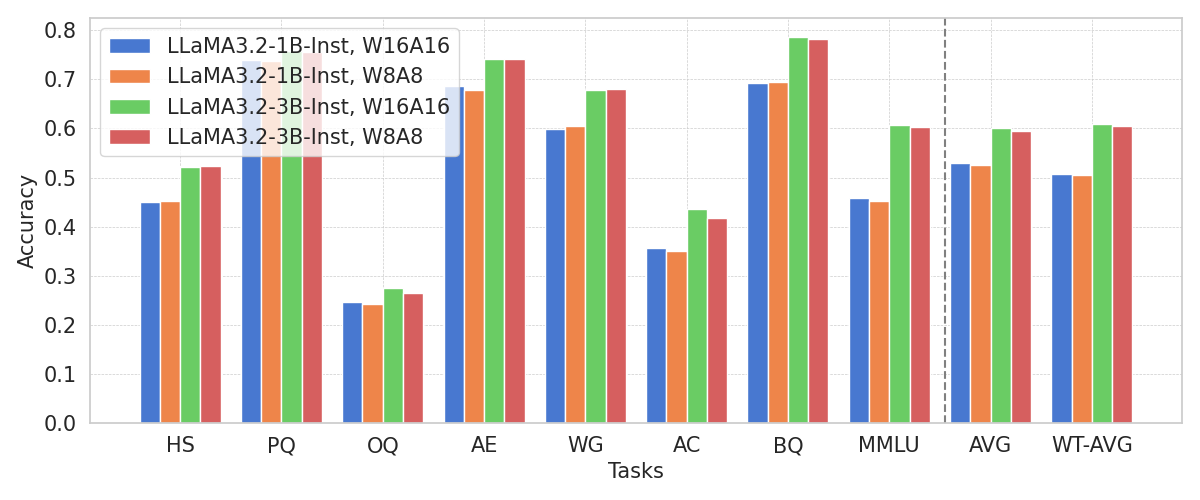}
    \caption{LLaMA3.2 models with W8A8 per-channel quantization}
    \label{fig:llama_3.2}
\end{figure}
\begin{figure*}[h]
    \centering
    \includegraphics[width=1\linewidth]{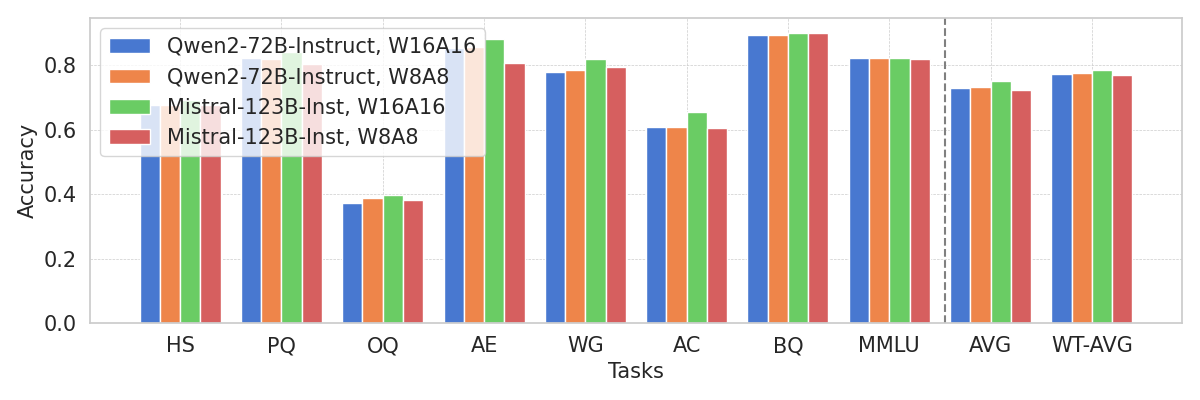}
    \caption{Accuracy of other Transformer-based models with W8A8 per-channel quantization}
    \label{fig:qwen_mistral_model_acc}
\end{figure*}
\begin{figure*}[h]
    \centering
    \includegraphics[width=1\linewidth]{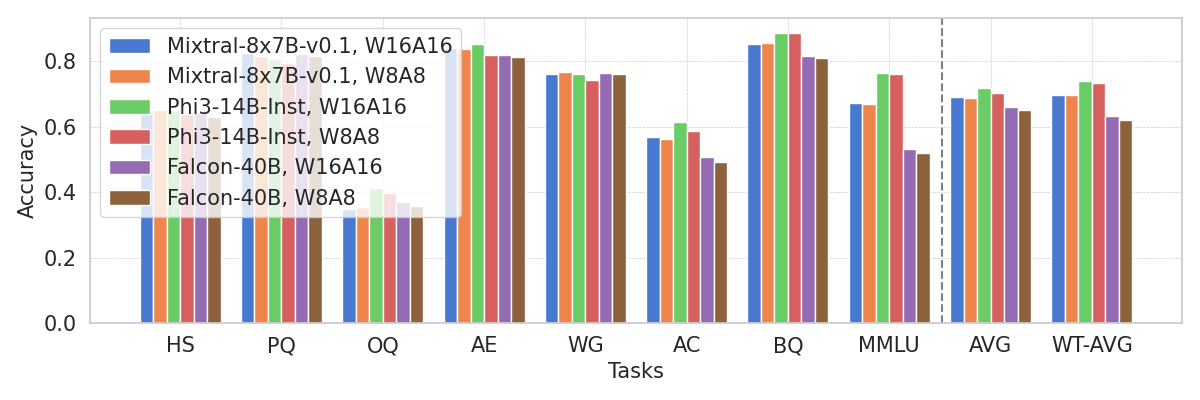}
    \caption{Accuracy of other Transformer-based models with W8A8 per-channel quantization}
    \label{fig:mixtral_phi_falcon_model_acc}
\end{figure*}

\subsection{LLaMA3-70B  is the ONLY ONE sensitive to W8A8 }
Figure~\ref{fig:llama_else} , Figure~\ref{fig:qwen_mistral_model_acc} and Figure~\ref{fig:mixtral_phi_falcon_model_acc} compare the accuracy of other models with different architectures with W8A8  quantization. 
\claude{The analysis spans the LLaMA family, including LLaMA3-8B, LLaMA2-70B, and LLaMA2-70B-chat, as well as other prominent Transformer-based designs such as Qwen2-72B, Mixtral-8x7B, Mistral-123B, Phi3-13B, and Falcon-40B. These architectures represent the top-performing architectures in the LLM Open Leaderboard.}

The data presented in these figures illustrates a  contrast to the behavior observed in the LLaMA3-70B series. Specifically, the models examined here exhibit remarkable resilience to W8A8 per-channel quantization. In the majority of instances, the performance of W8A8 quantized versions closely approximates that of their FP16 counterparts. 

\subsection{Why LLaMA3-70B is the only one?}

In order to understand the uniqueness of LLaMA3-70B models, we explore the weight distributions.

  \begin{figure}[h]
    \centering
    \includegraphics[width=0.7\linewidth]{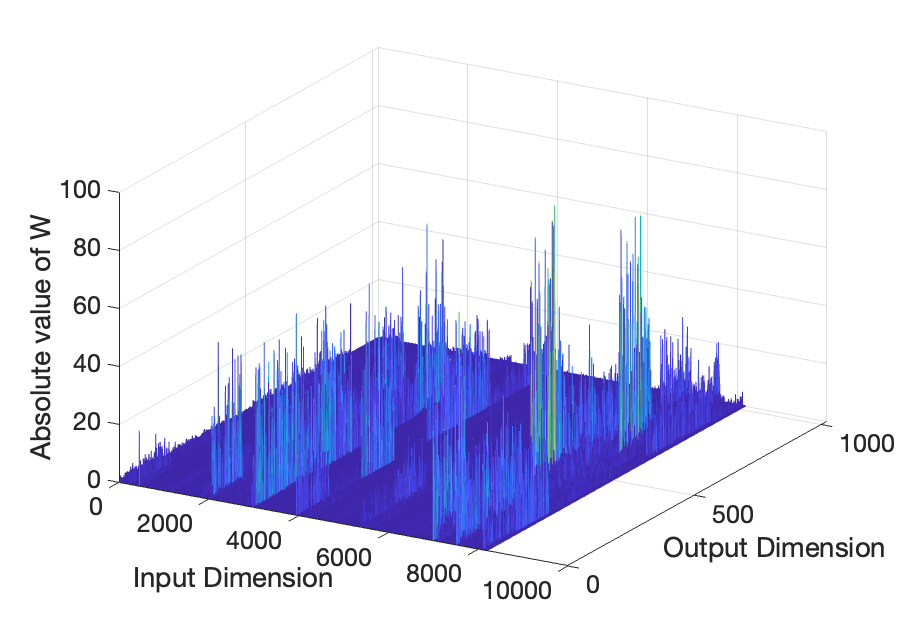}
    \caption{abs\_weights of the V matrix in the first Transformer block of LLaMA3.1-70B}
    \label{fig:wd-llama3.1-70b}
\end{figure}

\begin{figure}[htbp]
    \centering
    \includegraphics[width=0.7\linewidth]{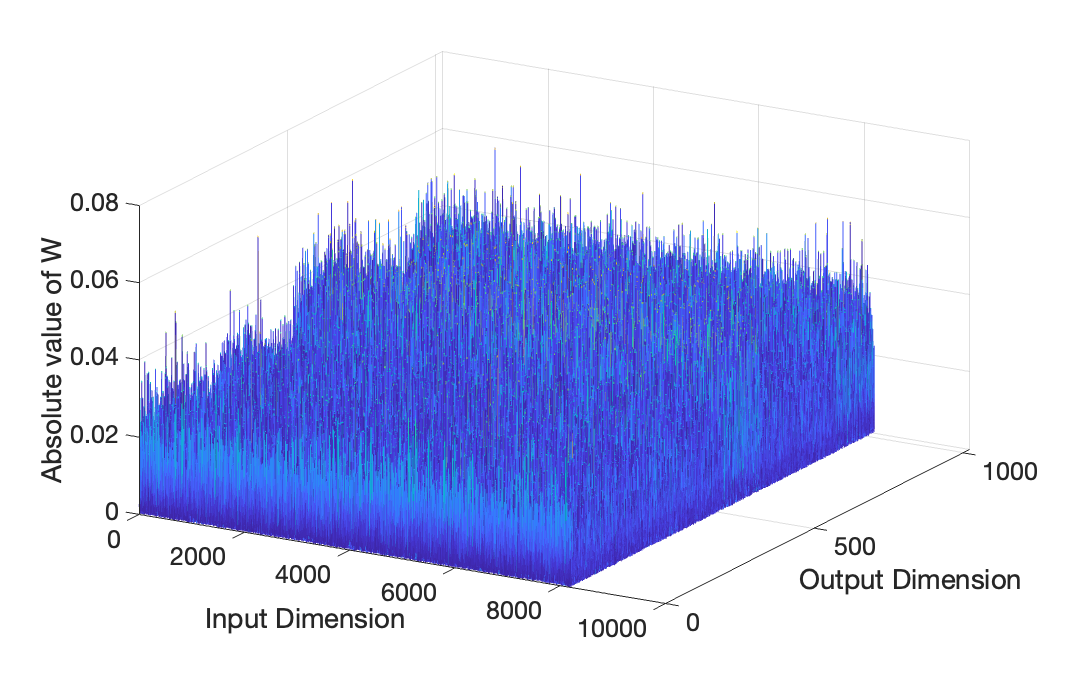}
    \caption{abs\_weights of the V matrix in the first Transformer block of LLaMA2-70B}
    \label{fig:wd-llama2-70b}
\end{figure}

\begin{figure}[h]
    \centering
    \includegraphics[width=0.7\linewidth]{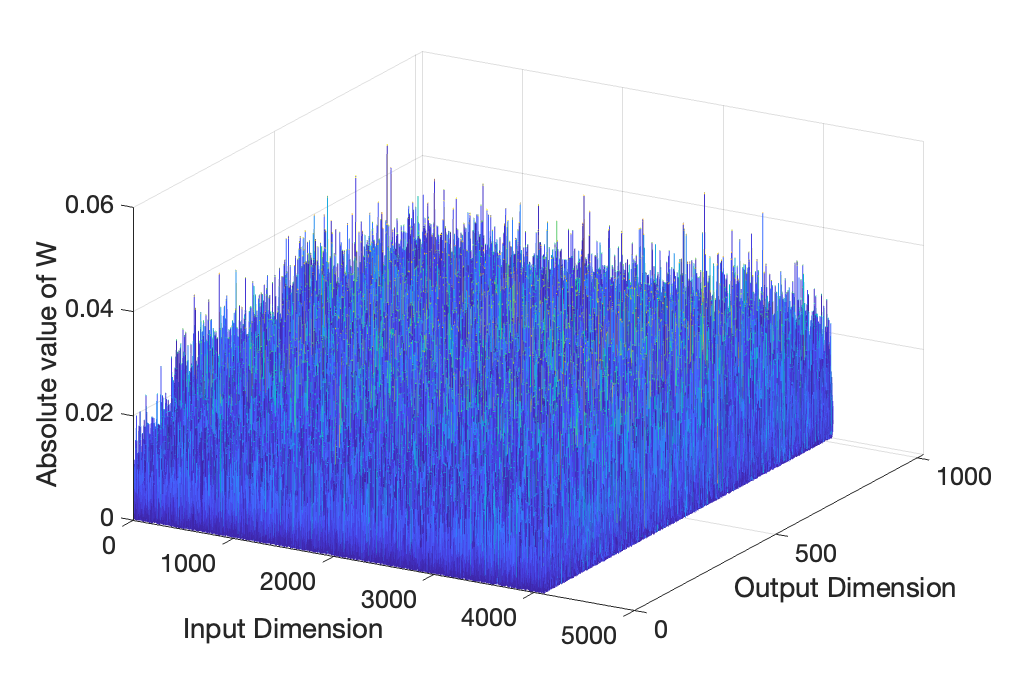}
    \caption{abs\_weights of the V matrix in the first Transformer block of LLaMA3-8B}
    \label{fig:wd-llama3-8b}
\end{figure}

\subsubsection{A qualitative exploration}

Figures~\ref{fig:wd_llama3-70b},~\ref{fig:wd-llama3.1-70b},~\ref{fig:wd-llama2-70b}, and~\ref{fig:wd-llama3-8b}, illustrate the absolute values of the V matrix weights in the first Transformer block for LLaMA3-70B, LLaMA3.1-70B, LLaMA3-8B, and LLaMA2-70B models, respectively. Other weight matrices for other models show similar trend and are shown in the Appendix. It is important to note that in per-channel quantization, each group comprises a length-$N$ vector along the input dimension. Our analysis reveals several key observations:

\begin{itemize} 
\item The maximum absolute weight values (max\_abs) in LLaMA3-70B ($93$) and LLaMA3.1-70B ($92.5$) surpass those in LLaMA3-8B ($0.05$) and LLaMA2-70B ($0.07$) by approximately three orders of magnitude. Furthermore, LLaMA3/3.1-70B exhibits a non-trivial number of weights with large magnitudes along the output dimension, which we term ``weight outliers''. This extensive range of weight values results in large quantization intervals for LLaMA3/3.1-70B, leading to substantial quantization errors in groups containing any weight outliers.
\item LLaMA3-70B displays a distinct pattern where weights with large max\_abs values cluster at specific input indices, forming visible "walls" in Figure~\ref{fig:wd_llama3-70b} and~\ref{fig:wd-llama3.1-70b}. In contrast, LLaMA2-70B and LLaMA3-8B lacks such obvious patterning.
\end{itemize}

\begin{figure}[h]
    \centering
    \includegraphics[width=0.8\linewidth]{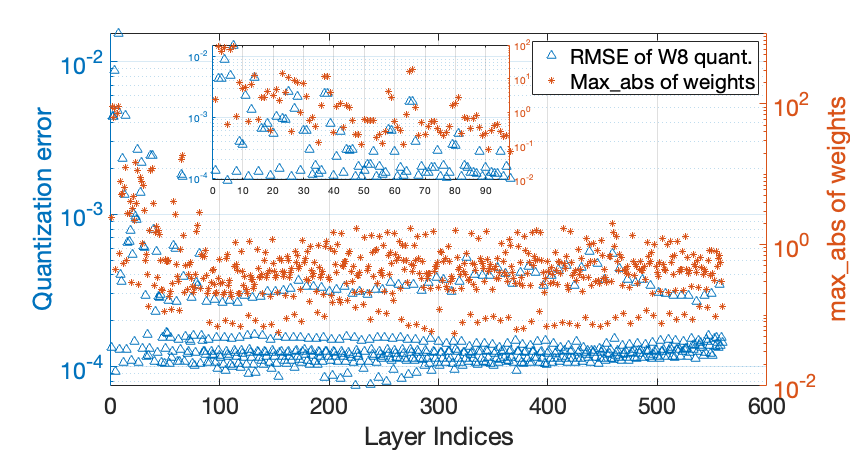}
    \caption{Quantization error and max\_abs of weights in each layer in LLaMA3-70B. LLaMA3-70B-Instruct, LLaMA3.1-70B, and LLaMA3.1-70B-Instruct have similar patterns.}
    \label{fig:qe_max_llama3-70b}
\end{figure}

\begin{figure}[htbp]
    \centering
    \includegraphics[width=0.8\linewidth]{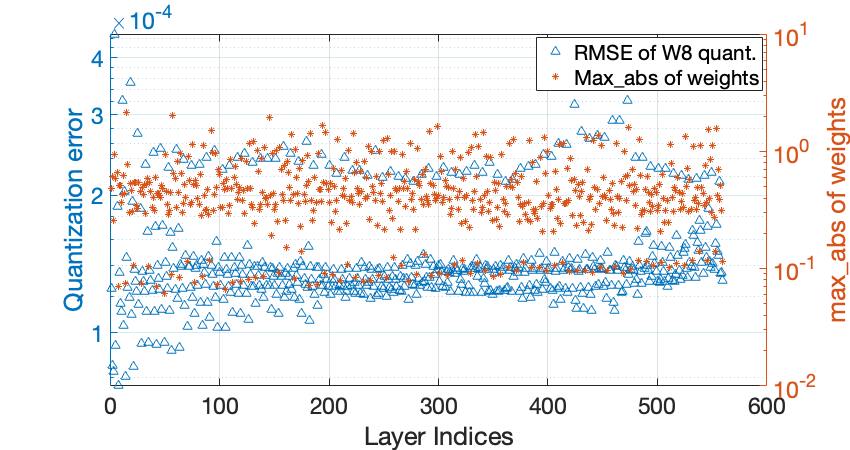}
    \caption{Quantization error and max\_abs of weights in each layer in LLaMA2-70B}
    \label{fig:qe_max_llama2-70b}
\end{figure}

\begin{figure}[h]
    \centering
    \includegraphics[width=0.8\linewidth]{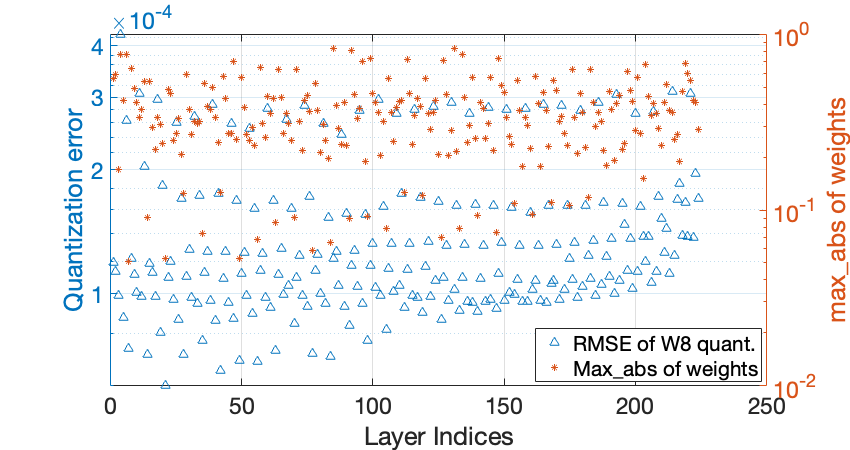}
    \caption{Quantization error and max\_abs of weights in each layer in LLaMA3-8B}
    \label{fig:qe_max_llama3-8b}

\end{figure}
\begin{figure}[h]
    \centering
    \includegraphics[width=0.8\linewidth]{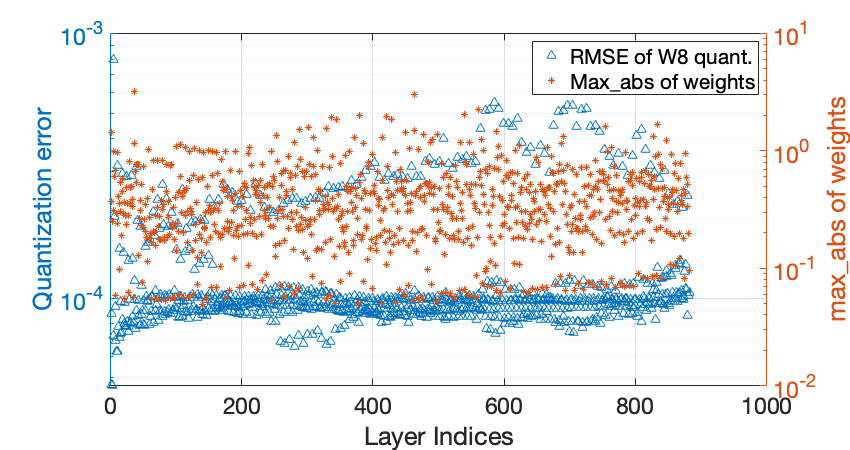}
    \caption{Quantization error and max\_abs of weights in each layer in LLaMA3.1-405B}
    \label{fig:qe_max_llama3-1-405b}
\end{figure}

\subsubsection{A quantitative exploration}

\claude{To quantitatively assess the robustness of weights under W8 quantization, we employ two key metrics. The first is the maximum absolute value (max\_abs) of weights for each layer in the model, which determines the scale factor and, consequently, the quantization interval. The second metric is the quantization error, measured by the root-mean-square error (RMSE) between the FP16 and 8-bit weights. These metrics are positively correlated, as larger quantization intervals typically result in greater errors.}

\claude{Figure~\ref{fig:qe_max_llama3-70b}, ~\ref{fig:qe_max_llama2-70b},~\ref{fig:qe_max_llama3-8b},~\ref{fig:qe_max_llama3-1-405b}  illustrate the max\_abs and quantization error for four LLaMA models: LLaMA3-70B, LLaMA2-70B, LLaMA3-8B, and the recently released LLaMA3.1-405B. The x-axis represents the total number of layers, excluding the initial embedding layer and the final output head. Each Transformer block in a LLaMA model comprises seven matrices (Q, K, V, O, up, gate, down), with 80, 80, 32, and 126 blocks in the four models, respectively. The left y-axis denotes the quantization error, while the right y-axis indicates the max\_abs weight of each layer.

}

Our comprehensive analysis of weight distributions across various layers and different LLM models yields the following conclusions:
\begin{itemize}
    \item In the initial layers, the LLaMA3-70B model exhibits weight outliers with max\_abs values that surpass those of its latter layers and all layers in other models by orders of magnitude. See the subfigure in Figure~\ref{fig:qe_max_llama3-70b} for a zoom-in view of the first 98 layers (14 blocks). This results in significantly larger quantization intervals, leading to quantization errors that are 1-2 orders of magnitude greater than in other models or layers.
    \item In the latter layers, the LLaMA3-70B model demonstrates behavior similar to other models in terms of quantization errors and max\_abs of weights. This observation suggests that the sensitivity of the LLaMA3-70B model series to W8 quantization may primarily stem from the initial layers. We will provide further confirmation of this hypothesis in the subsequent section.
\end{itemize}

\section{A Mixed Grouping Strategy}

\begin{figure*}[h]
    \centering
    \includegraphics[width=1\linewidth]{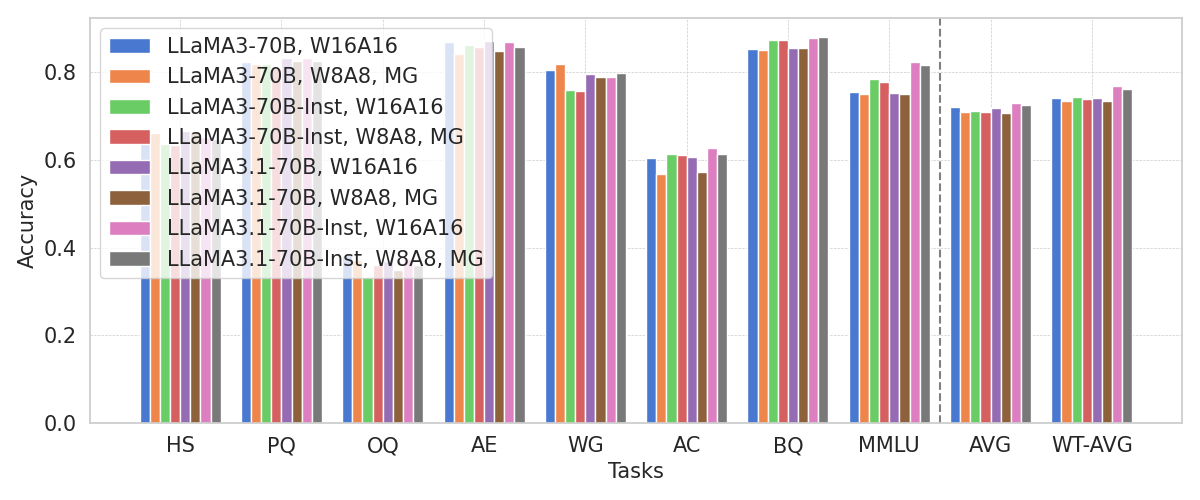}
    \caption{Accuracy comparison of  LLaMA3-70 model series with mixed grouping}
    \label{fig:llama3_70b_mixed_grouping}
\end{figure*}
\begin{figure*}[h]
    \centering
    \includegraphics[width=1\linewidth]{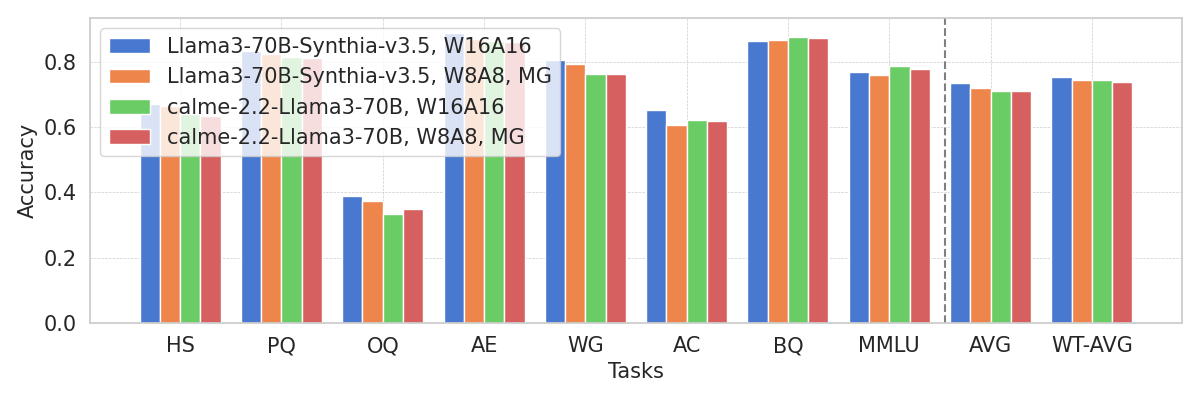}
    \caption{Accuracy comparison of  LLaMA3-70 derivative models with mixed grouping}
    \label{fig:llama3_70b_derivative_mixed_grouping}
\end{figure*}

Based on our analysis of Figure~\ref{fig:qe_max_llama3-70b} , we have empirically identified that in the LLaMA3-70B model, the Q, K, V, Up, and Gate matrices of Block 0, 1, and 3 exhibit exceptionally high quantization errors and maximum absolute weight values. To mitigate these quantization errors, we propose implementing a finer grouping granularity specifically for these 15 layers.

It is noteworthy that the layers requiring per-group quantization, as opposed to per-channel quantization, constitute only 2.68\% (15/560) of the total layers in the model. While many quantization studies employ group sizes of 128 or smaller (e.g., GPTQ, AWQ), our investigation reveals that a group size of 1024 is sufficient to prevent accuracy degradation while maintaining a larger group size for improved hardware efficiency.

\subsection{Main experimental results on mixed grouping}
Figure~\ref{fig:llama3_70b_mixed_grouping}  and Figure~\ref{fig:llama3_70b_derivative_mixed_grouping} presents a comparative analysis of the accuracy achieved by our proposed method against that of FP16 models. The presented models includes several LLaMA3-70 and its variants and derivatives in the Open LLM Leaderboard. The results convincingly demonstrate that our mixed quantization approach—combining 2.68\% per-group and 97.32\% per-channel 8-bit quantization—can match the accuracy of the LLaMA3-70B model with FP16 precision.


This improvement can be attributed to the finer granularity of the quantization process in these critical layers. By reducing the group size from the entire channel to 1024 elements, we effectively create multiple quantization groups within each channel. This approach allows for more precise representation of weight values, particularly in the presence of outliers that previously dominated the quantization scale factors.

In summary, this hybrid quantization strategy effectively addresses the unique challenges posed by the LLaMA3-70B model's weight distribution, particularly in its initial layers, while minimizing the hardware complexity overhead associated with per-group quantization across the entire model.

\begin{figure}[h]
    \centering
    \includegraphics[width=0.9\linewidth]{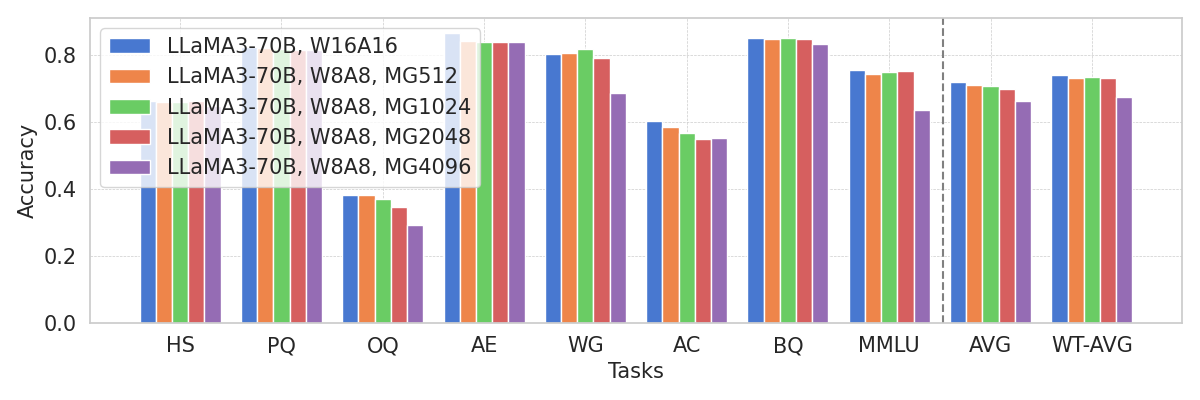}
    \caption{Ablation study on the mixed group size for Q, K, V, Up, and Gate of Block 0, 1, 3 in LLaMA3-70B}
    \label{fig:llama3_70b_MG_ablation}
\end{figure}

\subsection{Ablation on the group size}
We examine the influence of varying group sizes on the accuracy of W8A8 quantization for the LLaMA3-70B model. Figure~\ref{fig:llama3_70b_MG_ablation} illustrates this comparative analysis. Our results reveal that certain datasets, notably AE, AC, and OQ, exhibit particular sensitivity to increased group size applied exclusively to a total of 15 layers within Blocks 0, 1, and 3. From a holistic perspective, per-group quantization employing sizes of 512 or 1024 emerges as a good compromise between model accuracy and hardware efficiency.

\newtheorem{theorem}{Theorem}[section]

\section{A Bi-smoothing Strategy}

In~\cite{xiao2024smoothquantaccurateefficientposttraining}, it was proposed to mitigate activation outliers by reducing the maximum absolute values of activations while amplifying the maximum absolute values of weights. This approach assumes that activation outliers are concentrated in certain channels and are significantly more severe than weight outliers. However, as demonstrated in previous sections, this assumption no longer holds for the LLaMA3-70B and LLaMA3.1-70B model series, including their instruction-tuned versions. In this section, we introduce bi-directional smoothing algorithms that balance the magnitudes of both weights and activations.

\subsection{Equivalence in bi-directional smoothing}
Let $W\in \mathbb{R}^{M\times N}$ be the weights and $A \in \mathbb{R}^{B\times L\times N}$ be the input activation. In the representation of a Transformer linear layer, $M$ is output dimension, $N$ is the hidden dimension, $B$ is the batch size and $L$ is the sequence length. The output of the linear layer $Y \in \mathbb{R}^{B\times L\times M}$ is calculated by
\[
Y[i,:,:] = A[i,:,:] W^T, \forall 0\leq i \leq B-1,
\]
where we use Python tensor representation for the 3D tensor $A$, $Y$ and 2D tensor $W$.

There is an equivalence when multiplying $A[i,:,:]$ and $W$. 
\begin{theorem}
Let $S\in \mathbb{R}^{N}$ be a non-zero vector. We denote $W_s$ and $A_s$ the column-wise (last dimension) smoothed tensors, respectively, such that
$$
W_s[j,k] = W[j,k] S[k],  \forall 0\leq j \leq M, 0 \leq k \leq N-1,
$$
and
$$
A_s[i,j,k] = A[i,j,k] / S[k], \forall 0\leq i\leq B-1, 0\leq j\leq L-1,  0 \leq k \leq N-1,
$$
Then the following equation holds,
\[
A[i,:,:] W^T = A_s[i,:,:] W_s^T , \forall 0\leq i \leq B-1.
\]
\end{theorem}
\begin{proof}
Let $Y[i,:,:]=A[i,:,:] W^T$ and $Y_s[i,:,:]=A_s[i,:,:] W_s^T$, then $ \forall 0\leq i\leq B-1, 0\leq j\leq L-1,  0 \leq k \leq M-1$,
\begin{align*}
    Y_s[i,j,k] &= \sum_z A_s[i,j,z] W_s[z,j] \\
               &= \sum_z  A[i,j,z] / S[z] * W[z,j] S[z] \\
               &= \sum_z A[i,j,z] W[z,j] \\
               & = Y[i,j,k].
\end{align*}
\end{proof}
We call the vector $S\in \mathbb{R}^{N}$ the \textbf{smooth-factor} for $A$ and $W$.

\subsection{How to select the smooth-factor?}
For per-channel quantization, a quantization group consists of $N$ numbers and the max\_abs in the quantization group determines the scale-factor and quantization interval. First, assume the batch size is 1, and then $A\in \mathbb{R}^{1\times L \times N}$. Since it is desirable to minimize the max\_abs of both weight  and input activation, but the multiplication of these two values, i.e., $\max \left( \lvert W[:,i] \rvert \right)$ and $\max \left( \lvert A[0,:,i]\rvert \right)$, remains constant to guarantee the output of $A W^T$ is unchanged. Therefore, we choose the smooth-factor as
\begin{align*}
    S[k] = \sqrt{\frac{\max_j \left( \lvert A[0,j,k]\rvert \right)}{\max_j \left( \lvert W[j,k] \rvert \right)}}, \forall 0 \leq k \leq N-1.
\end{align*} 
Then we have
\begin{align*}
     \max_j \left( \lvert W_s[j,k] \rvert \right)&=  \max_j \left( \lvert W[j,k] S[i] \rvert \right) \\
     &=\sqrt{\max_j \left( \lvert A[0,j,k]\rvert \right) \max_j\left( \lvert W[j,k] \rvert \right)} \\
     &= \max_j \left( \lvert A[0,j,k] S[k]\rvert \right) \\
     & = \max_j \left( \lvert A_s[0,j,k] \rvert \right),
\end{align*}
i.e., the max\_abs after the bi-smoothing for weight and input activation is balanced. 

In general when the batch size $B$ is greater than 1, then we can choose $S[k]$ as
  \begin{align}  \label{eq:smooth-factor}
    S[k] = \sqrt{\frac{\textrm{median}_i \max_j \left( \lvert A[i,j,k]\rvert \right)}{\max_j \left( \lvert W[j,k] \rvert \right)}}, \forall 0 \leq k \leq N-1.
\end{align}

\subsection{Main experiment results for bi-smoothing}

We show the accuracy comparison of all four LLaMA3-70B series of models (LLaMA3.1-70B-Instruct, LLaMA3.1-70B, LLaMA3-70B-Instruct, and LLaMA3-70B) in the section.
\begin{figure}
    \centering
    \includegraphics[width=\linewidth]{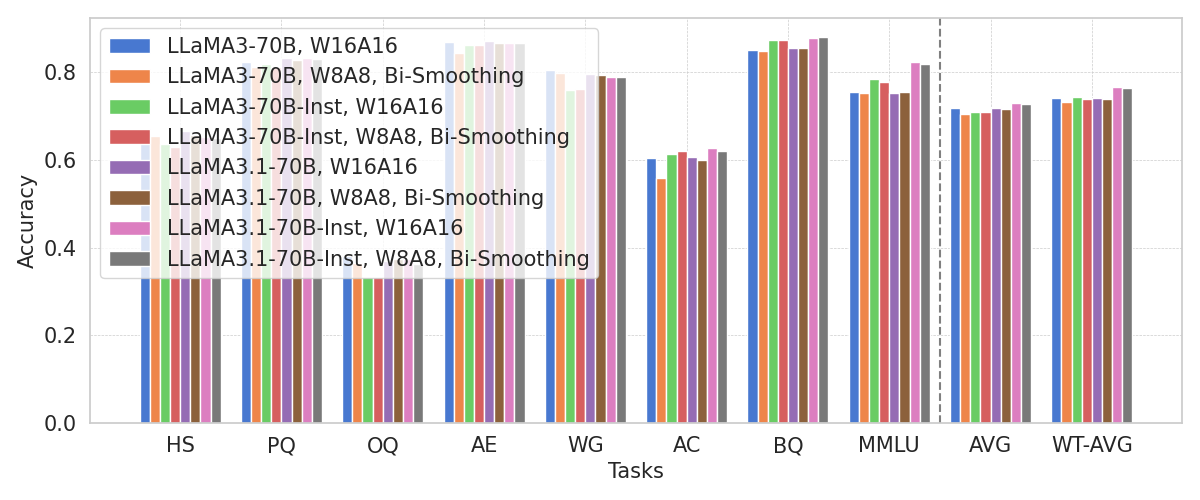}
    \caption{Accuracy comparison of LLaMA3-70B model series with bi-smoothing}
    \label{fig:llama3.1-70b-bismooth}
\end{figure}

Figure~\ref{fig:llama3.1-70b-bismooth} demonstrates the effectiveness of the proposed bi-smoothing method for the entire LLaMA3-70B model series. It is observed that the  accuracy  of  models with per-channel W8A8  quantization are as good as their FP16 counterparts, which shows significant improvement over W8A16 models in Figure~\ref{fig:llama3-70b-w8a8}.  Note that to calculate the smooth-factor in Equation~(\ref{eq:smooth-factor}), we need some calibration data to estimate $\textrm{median}_i \max_j \left( \lvert A[i,j,k]\rvert \right)$. To prevent data contamination, we use only eight questions from the GPQA reasoning task for estimation. The calibration process takes only a few seconds, requiring a single inference pass over these samples. As demonstrated in the later ablation studies, the accuracy of the bi-smoothing method remains stable across varying amounts and sources of calibration data.

\subsubsection{Why bi-smoothing works well? }
\begin{figure}[t]
    \centering
    \includegraphics[width=1\linewidth]{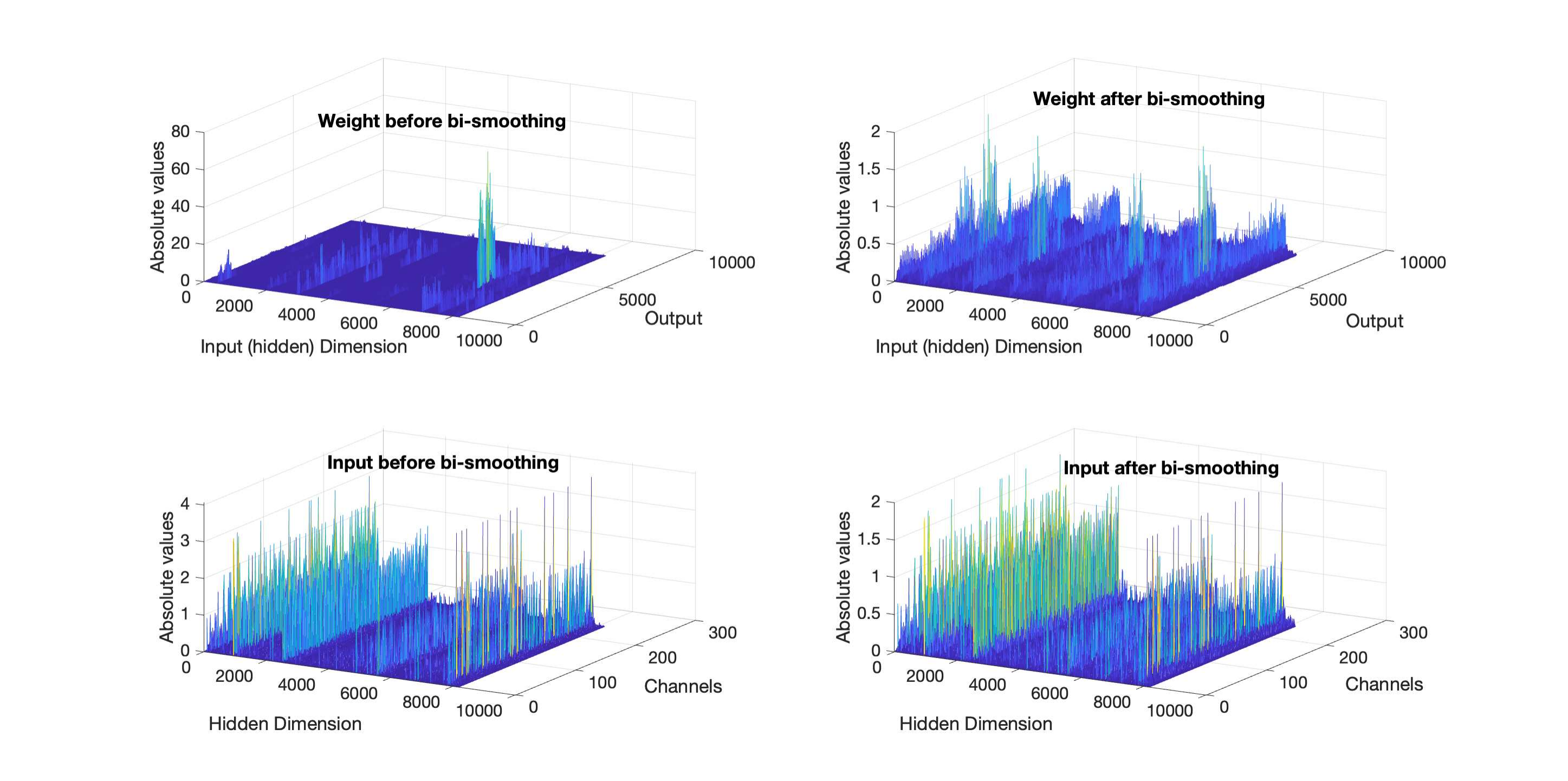}
    \caption{The plot illustrates the absolute values of weights and input activations in the LLaMA3.1-70B-Instruct model, specifically focusing on the first Q matrix in the initial Transformer block. The input activations are multiplied by the Q matrix. \textbf{Top-left}: Before bi-smoothing, large weight values (ranging from 60 to 80) form distinct "walls" parallel to the output axis. \textbf{Top-right}: After bi-smoothing, these weight "walls" remain but with significantly reduced magnitudes (below 2.0). \textbf{Bottom-left}: Input activations before bi-smoothing. \textbf{Bottom-right}: After bi-smoothing, the input activations are also suppressed, exhibiting smaller magnitudes.
}
    \label{fig:before-and-after-bi-smooth}
\end{figure}

Figure~\ref{fig:before-and-after-bi-smooth} compares the weights and input activations of the first Q matrix in the LLaMA3.1-70B-Instruct model. The input activations are drawn from a sequence of length 243 in the PIQA dataset. Notably, problematic weight outliers, with magnitudes exceeding 60, are suppressed to below 2.0, alongside activation outliers. This smoothing process makes both weights and activations far more resilient to quantization, achieving W8A8 accuracy nearly identical to that of the W16A16 configuration (as detailed in the next section).

\begin{figure} [h]
    \centering
    \includegraphics[width=0.8\linewidth]{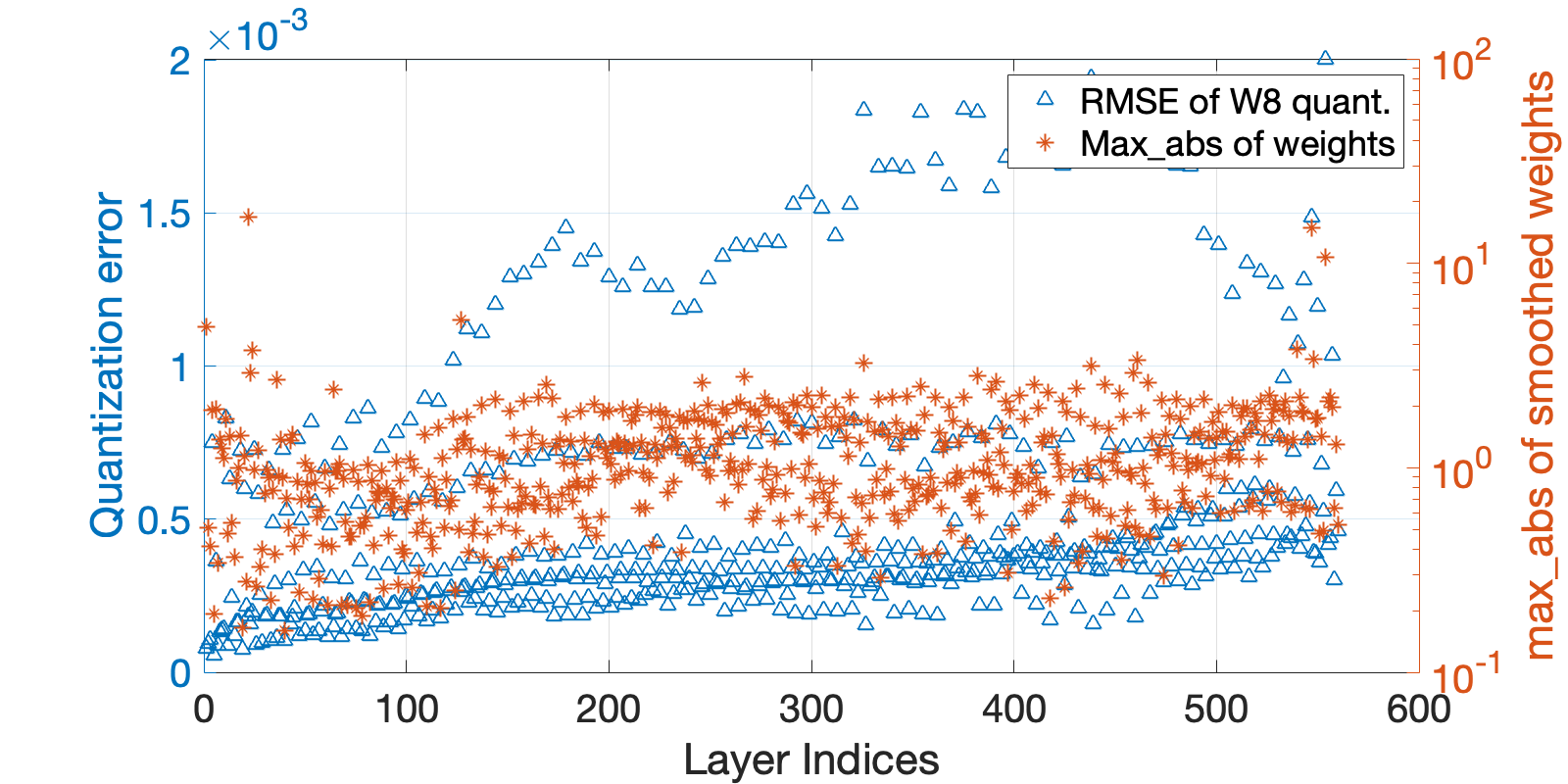}
    \caption{Quantization error and max\_abs of weights after bi-smoothing of each layer in the LLaMA3.1-70B-Instruct model.}
    \label{fig:qe_max_llama3-1-70b-inst-bi-smoothed}
\end{figure}

In general, Figure~\ref{fig:qe_max_llama3-1-70b-inst-bi-smoothed} illustrates the quantization error and max\_abs across all layers of the LLaMA3.1-70B-Instruct model after bi-smoothing. Compared to Figure~\ref{fig:qe_max_llama3-70b}, depicting the unsmoothed model, there is a substantial reduction in max\_abs, yielding an order-of-magnitude decrease in quantization error.

\subsection{Ablation study on bi-smoothing}

From Figure~\ref{fig:llama3.1-70b-bismooth}, we can see the best performing model is LLaMA3.1-70B-Instruct. So, we use this model for ablation study of bi-smoothing algorithms. We will show the bi-smoothing method is stable in calibration data and the choice of computing the smooth-factor.

\subsubsection{Calibration data source}

\begin{figure}
    \centering
    \includegraphics[width=1\linewidth]{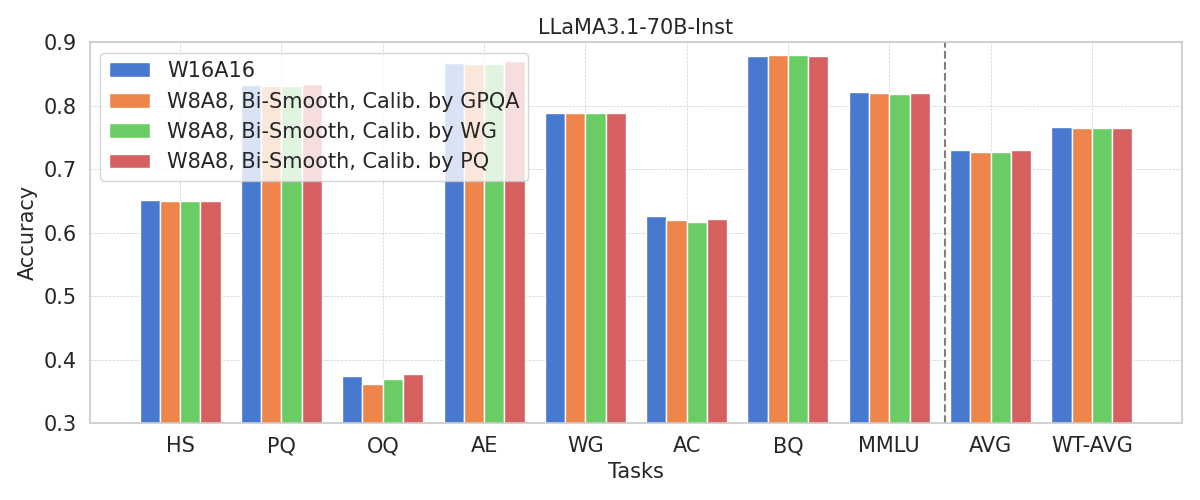}
    \caption{Different calibration data for bi-smoothing}
    \label{fig:calib-data-gpqa-wg-pq}
\end{figure}

Figure~\ref{fig:calib-data-gpqa-wg-pq} presents an accuracy comparison across different calibration data sources. We tested 8 samples from the GPQA dataset, 32 samples from the WG dataset, and 32 samples from the PQ dataset. The results indicate that the choice of calibration data has minimal impact on accuracy; all three datasets enable the W8A8 per-channel quantized model to achieve accuracy comparable to that of the FP16 counterpart.

\subsubsection{Calibration data size}

\begin{figure}[h]
    \centering
    \includegraphics[width=1\linewidth]{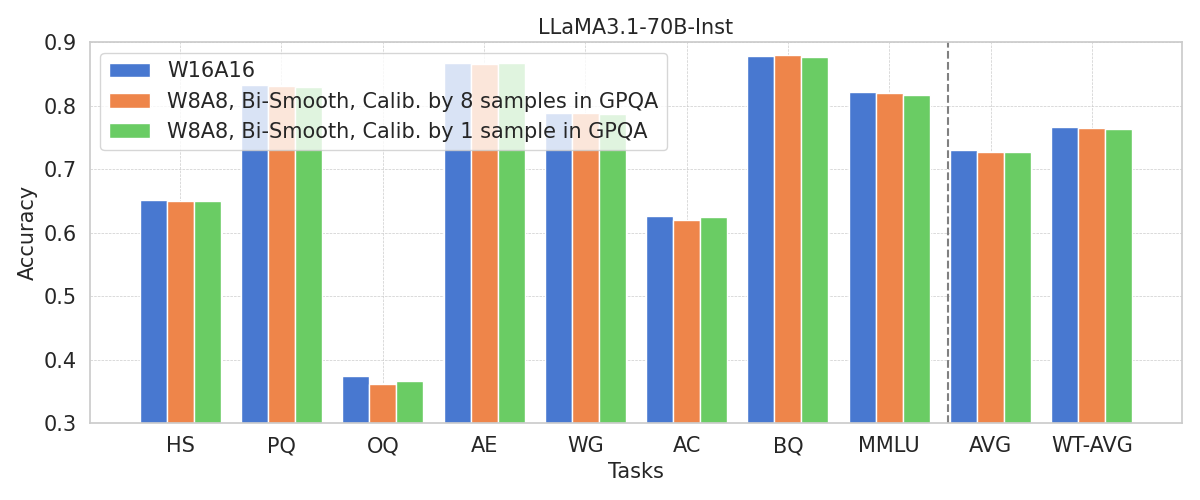}
    \caption{Comparison of calibration data using 8 samples and 1 sample from GPQA.}
    \label{fig:ablation_num_samples}
\end{figure}

We evaluated the impact of calibration data volume on smooth-factor computation, comparing results from 8 samples versus a single sample from the GPQA dataset. Figure~\ref{fig:ablation_num_samples} demonstrates that the difference in performance is negligible; both resulting W8A8 models achieve accuracy comparable to their FP16 counterpart.

\subsubsection{Smooth-factor calculation}

Note that in Equation (\ref{eq:smooth-factor}), we use ``median'' to calculate the smooth-factor. A natural alternative is to replace ``median'' by ``maximum''  in the following equation.
  \begin{align}  \label{eq:smooth-factor-alternative}
    S[k] = \sqrt{\frac{\max_i \max_j \left( \lvert A[i,j,k]\rvert \right)}{\max_j \left( \lvert W[j,k] \rvert \right)}}, \forall 0 \leq k \leq N-1.
\end{align}

\begin{figure}
    \centering
    \includegraphics[width=1\linewidth]{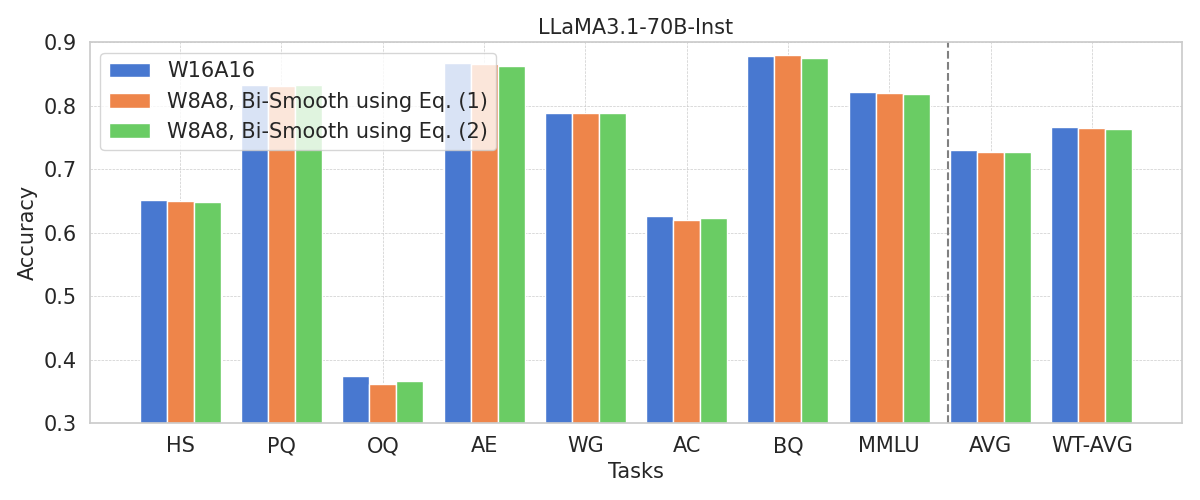}
    \caption{Comparison of using Eq.~\ref{eq:smooth-factor} and Eq.~\ref{eq:smooth-factor-alternative} to calculate the smooth-factor}
    \label{fig:ablation-smooth-factor-calculation}
\end{figure}

Figure~\ref{fig:ablation-smooth-factor-calculation}  illustrates the comparison, revealing that there is negligible difference between using the ``median'' or ``max'' to calculate the smooth factor. This finding underscores the robustness of the bi-smoothing method in recovering the accuracy of W8A8 models.

\subsubsection{Applying bi-smoothing to a subset of Transformer layers}

\begin{figure}
    \centering
    \includegraphics[width=1\linewidth]{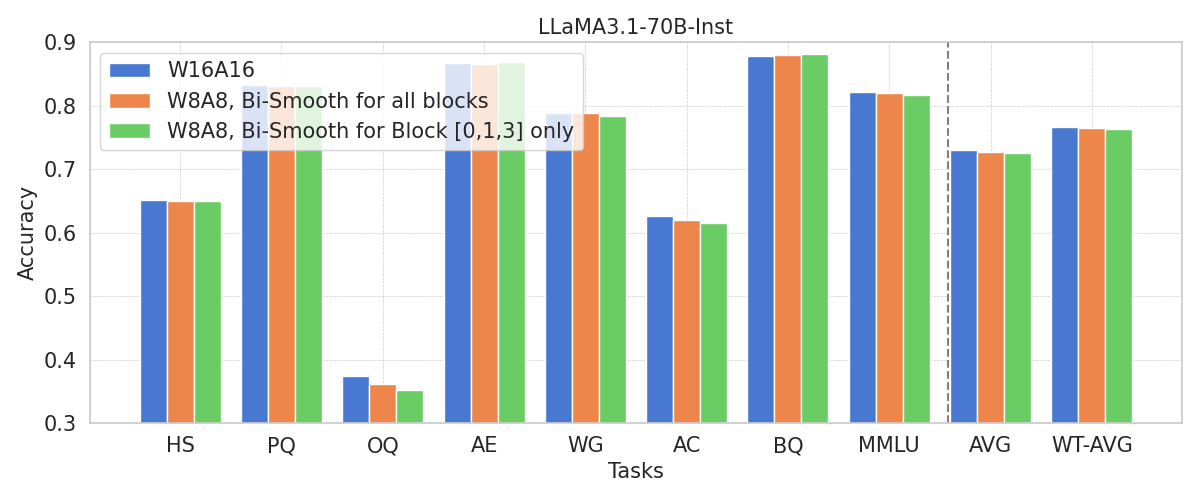}
    \caption{ Applying bi-smoothing methods to a subset of Transformer blocks}
    \label{fig:ablation_subblock}
\end{figure}

Our previous analysis revealed that significant weight outliers are predominantly confined to the initial Transformer blocks. An ablation study on the number of blocks with bi-smoothing method was conducted, with results illustrated in Figure~\ref{fig:ablation_subblock}. Notably, applying bi-smoothing to only the 0th, 1st, and 3rd Transformer blocks yields W8A8 accuracy nearly equivalent to that of the full-precision (FP16) model. This finding corroborates our hypothesis that the initial blocks are the primary source of the LLaMA3-70B model series' susceptibility to W8A8 quantization.

\section{Discussions}

According to Meta's technical report~\cite{dubey2024llama3herdmodels}, LLaMA2-70B and LLaMA3-70B share similar model architectures, while LLaMA3-8B/70B and LLaMA3.1-8B/70B/405B are reported to have similar training data and strategies. However, our analysis reveals significantly different weight distribution characteristics in LLaMA3/3.1-70B compared to other LLaMA models. This unexpected divergence in weight patterns, particularly the presence of numerous large outliers exclusively in LLaMA3/3.1-70B models, presents an intriguing peculiarity.

The origins of this phenomenon remain unclear, given that models sharing either architectural design or training methodologies do not exhibit similar characteristics. Unraveling the underlying mechanisms responsible for this unique weight distribution could potentially offer valuable insights into the learning processes of Large Language Models (LLMs), especially in the context of reasoning tasks.

\section{Related Works}
Quantization is useful in reducing model size and potentially increases model inference speed~\cite{han2016deepcompressioncompressingdeep,jacob2017quantizationtrainingneuralnetworks,nagel2019datafreequantizationweightequalization,bondarenko2021understandingovercomingchallengesefficient,wang2019haqhardwareawareautomatedquantization,bengio2013estimatingpropagatinggradientsstochastic}. There two main categories of quantization, post-training quantization (PTQ) and quantization-aware training (QAT).
\subsection{Post-training quantization}
PTQ does not use any back-propagation to finetune weights. For \textbf{weight-only quantization}, LLM.int8()~\cite{dettmers2022llmint8} uses a mixed precision with 16-bit MAC for outliers. GPTQ~\cite{frantar2023gptq} and AWQ~\cite{lin2023awq} quantize the per-group weight-only tensor using as small as 3-4 bits. SqueezeLLM~\cite{kim2024squeezellmdenseandsparsequantization} uses non-uniform quantization and store outliers efficiently. For~\textbf{weight-activation quantization} that enables faster inference~\cite{dettmers2022gptint,wei2023outliersuppressionpushinglimit} 
SmoothQuant~\cite{xiao2024smoothquantaccurateefficientposttraining} transfers the outliers from activation to weights to enable per-tensor W8A8. 
QServe~\cite{lin2024qservew4a8kv4quantizationcodesign} proposes W4A8K4 to serve LLM models.
Some other studies enable even smaller number of bits~\cite{shao2024omniquantomnidirectionallycalibratedquantization, zhao2024atomlowbitquantizationefficient, ashkboos2024quarotoutlierfree4bitinference}. 
Notably, our observations on the weights and activations of the LLaMA3-70B model series contrast with those of SmoothQuant. The proposed bi-smoothing method extends this approach by addressing both weight and activation outliers. Compared to other PTQ methods, we aim to maintain per-channel  instead of per-group quantization for maximal hardware efficiency.

\subsection{Quantization aware training}
Quantization-aware training (QAT) requires both calibration data and  back-propagation to finetune the quantize weights~\cite{bengio2013estimatingpropagatinggradientsstochastic,gholami2021surveyquantizationmethodsefficient}. While full finetuning~\cite{alpaca} can recover the accuracy to the maximum extent, Q-LoRA~\cite{dettmers2023qloraefficientfinetuningquantized} reduces the memory requirement significantly.

\section{Conclusions}
Our investigation reveals a distinctive vulnerability in the LLaMA3-70B model series, including LLaMA3.1-70B, to per-channel quantization. We attribute this susceptibility to significant weight outliers in the initial layers. The contrast in behavior between the LLaMA3-70B and other LLaMA models (LLaMA2, LLaMA3/3.1-8B, LLaMA3.1-405B), as well as other tested LLMs with different architectures, warrants further exploration of the underlying training strategies. To address this issue, we propose a mixed-grouping method that substantially enhances the accuracy of LLaMA3-70B from 45.4\% to 73.4\%, effectively matching the performance of its FP16 counterpart.


\clearpage

\section{Appendix}

\subsection{Models}
Table~\ref{tab:test_models} shows all the tested models downloaded from Huggingface repository.

\begin{table*}[h]
    \centering
    \begin{tabular}{l|l}
    \toprule
         Model name&  Huggingface URL\\ \hline
         LLaMA3-70B&  https://huggingface.co/meta-llama/Meta-Llama-3-70B\\
         LLaMA3-70B-Instruct&  https://huggingface.co/meta-llama/Meta-Llama-3-70B-Instruct\\
         LLaMA3.1-70B&  https://huggingface.co/meta-llama/Meta-Llama-3.1-70B\\
 LLaMA3.1-70B-Instruct&https://huggingface.co/meta-llama/Llama-3.1-70B-Instruct\\
         Llama3-70B-Synthia & https://huggingface.co/migtissera/Llama-3-70B-Synthia-v3.5 \\
         calme-2.2-llama3-70b  & https://huggingface.co/MaziyarPanahi/calme-2.2-llama3-70b \\
         LLaMA3-8B&  https://huggingface.co/meta-llama/Meta-Llama-3-8B\\
 LLaMA3.2-1B-Instruct&https://huggingface.co/meta-llama/Llama-3.2-1B-Instruct\\
 LLaMA3.2-3B-Instruct&https://huggingface.co/meta-llama/Llama-3.2-3B-Instruct\\
         LLaMA2-70B&  https://huggingface.co/meta-llama/Llama-2-70b-hf\\
         LLaMA2-70B-chat&  https://huggingface.co/meta-llama/Llama-2-70b-chat-hf\\
         Qwen2-72B & https://huggingface.co/Qwen/Qwen2-72B \\
         Mixtral-8x7B & https://huggingface.co/mistralai/Mixtral-8x7B-v0.1 \\
         Phi3-14B-Instruct &  https://huggingface.co/microsoft/Phi-3-medium-128k-instruct \\
         Mistral-Large-Instruct-123B& https://huggingface.co/mistralai/Mistral-Large-Instruct-2407 \\
         Falcon-40B & https://huggingface.co/tiiuae/falcon-40b \\
         \bottomrule
    \end{tabular}
    
    \caption{Tested models}
    \label{tab:test_models}
\end{table*}

\subsection{Weights of LLaMA3-70B}

\subsubsection{The first block has more large outliers}

\begin{figure}[h]
    \centering
    \includegraphics[width=1\linewidth]{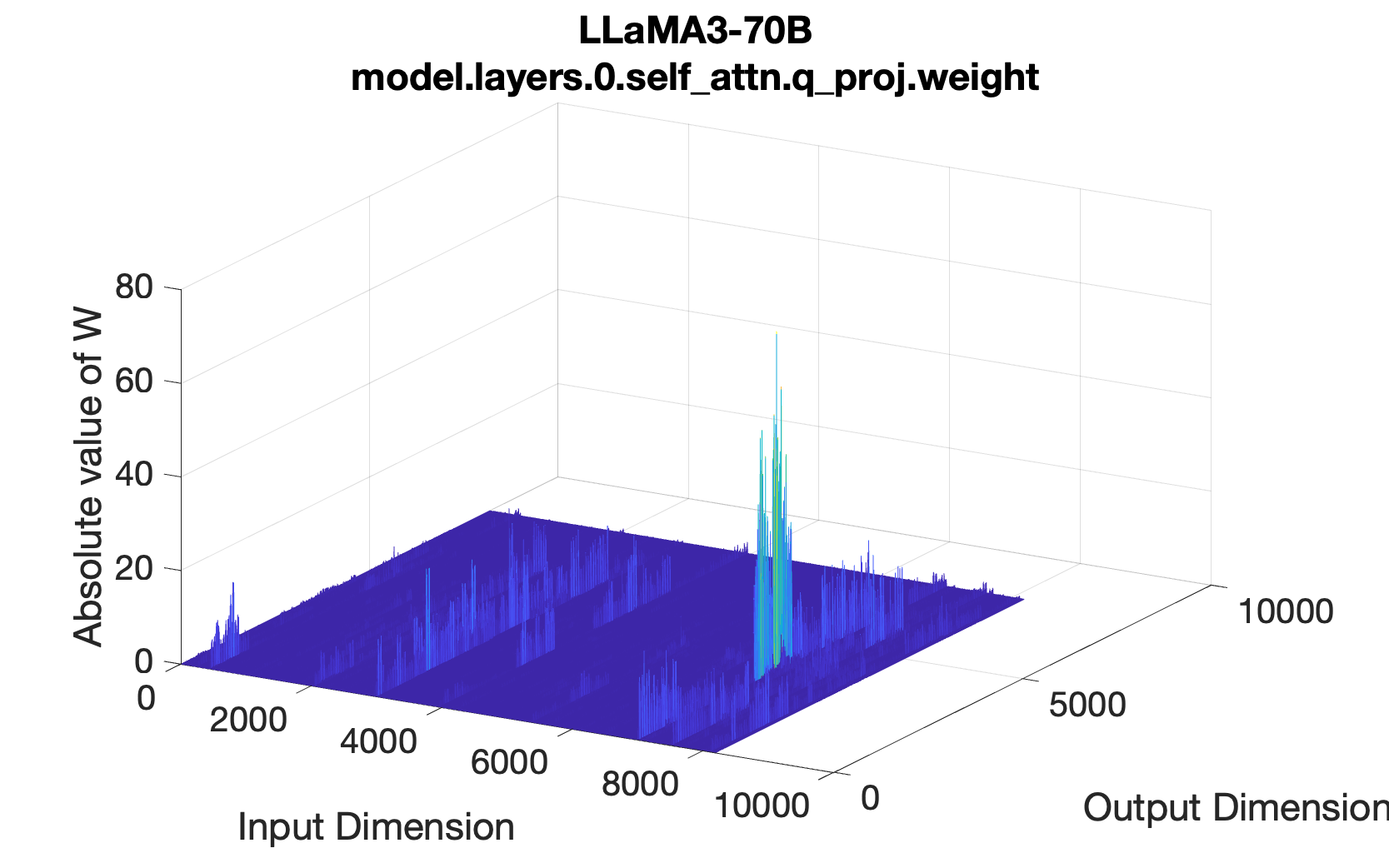}
    \caption{Weights in LLaMA3-70B }
    \label{fig:w_0_q}
\end{figure}

\begin{figure}[h]
    \centering
    \includegraphics[width=1\linewidth]{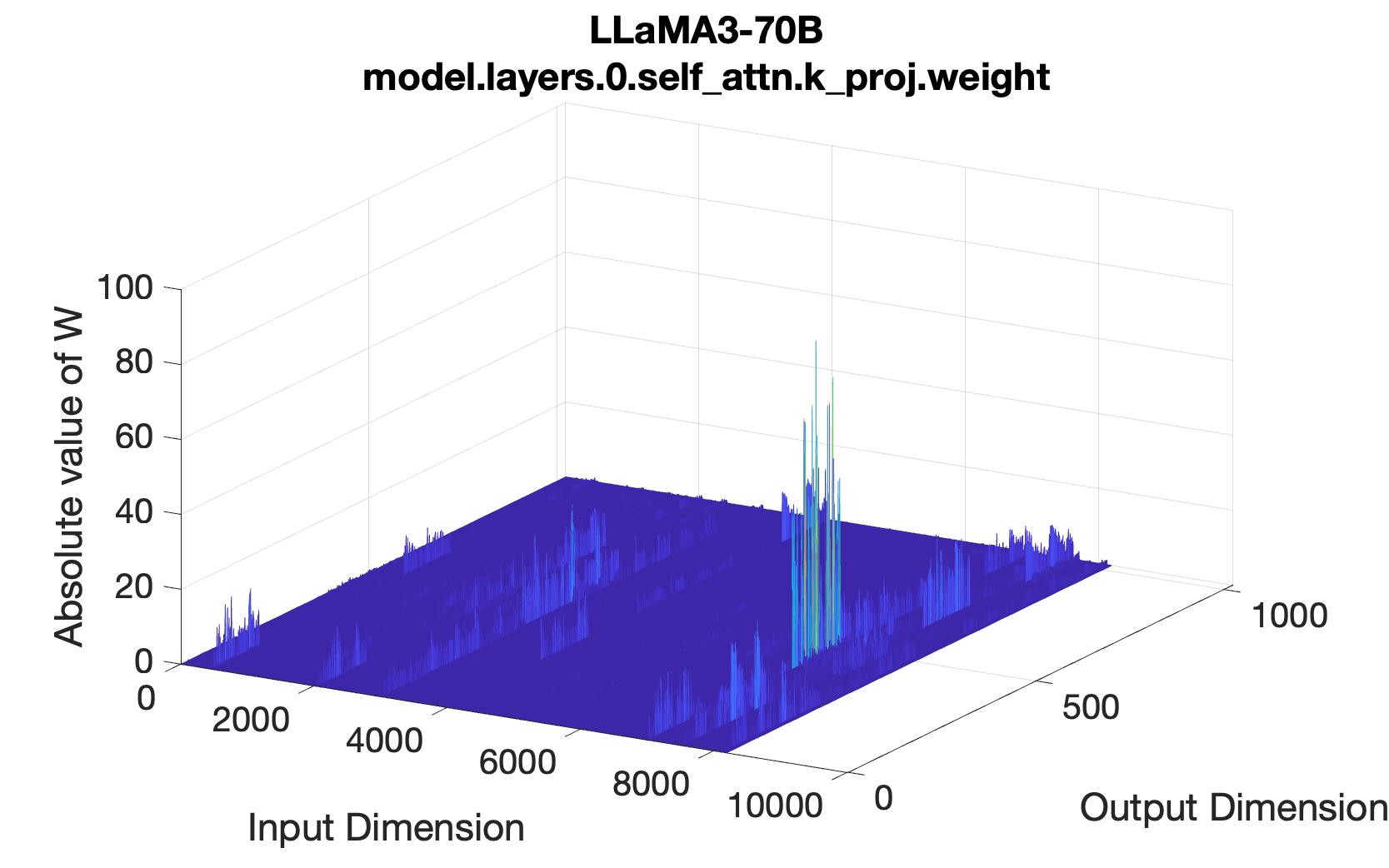}
    \caption{Weights in LLaMA3-70B }
    \label{fig:w_0_k}
\end{figure}

\begin{figure}[h]
    \centering
    \includegraphics[width=1\linewidth]{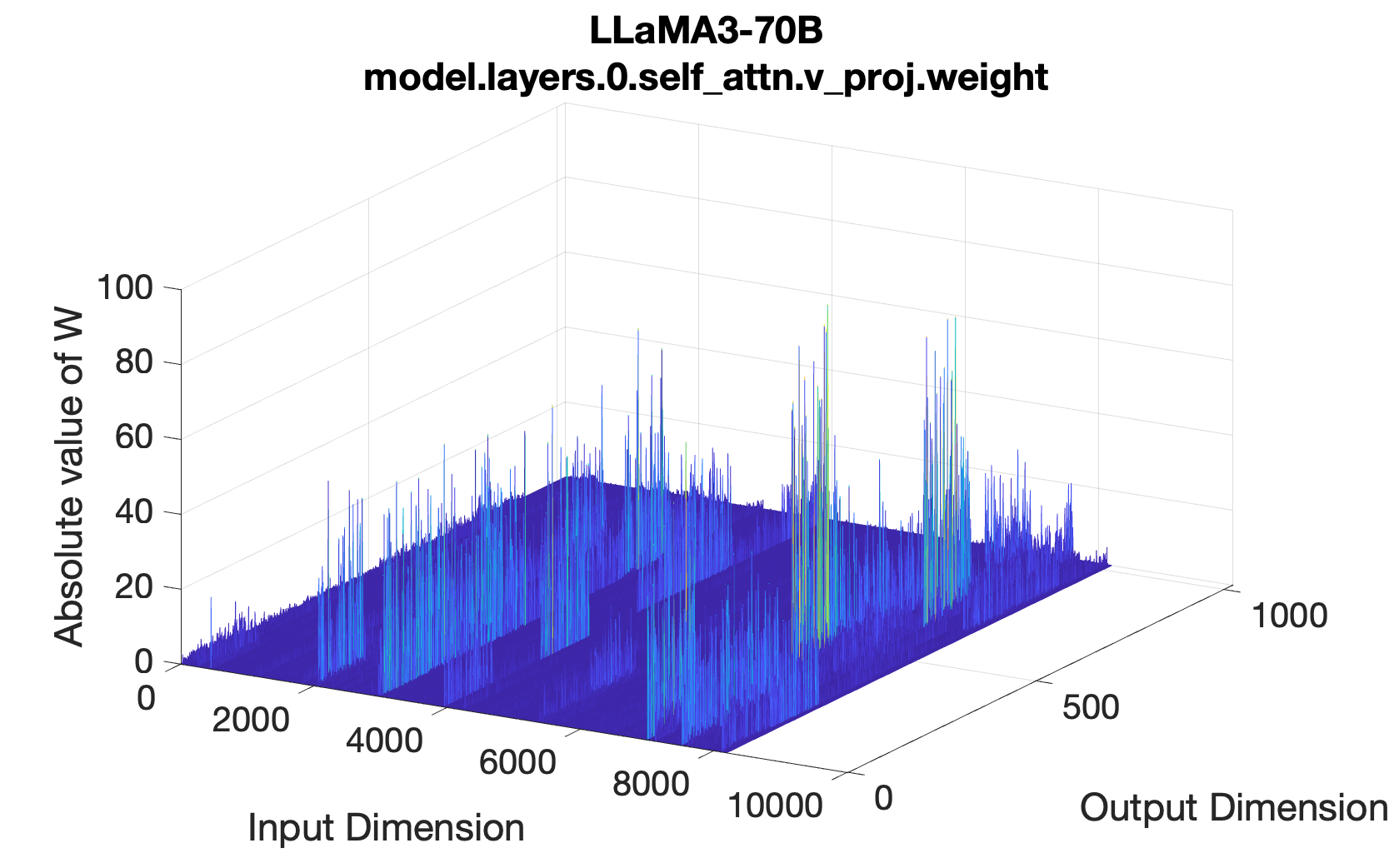}
    \caption{Weights in LLaMA3-70B }
    \label{fig:w_0_v}
\end{figure}

\begin{figure}[h]
    \centering
    \includegraphics[width=1\linewidth]{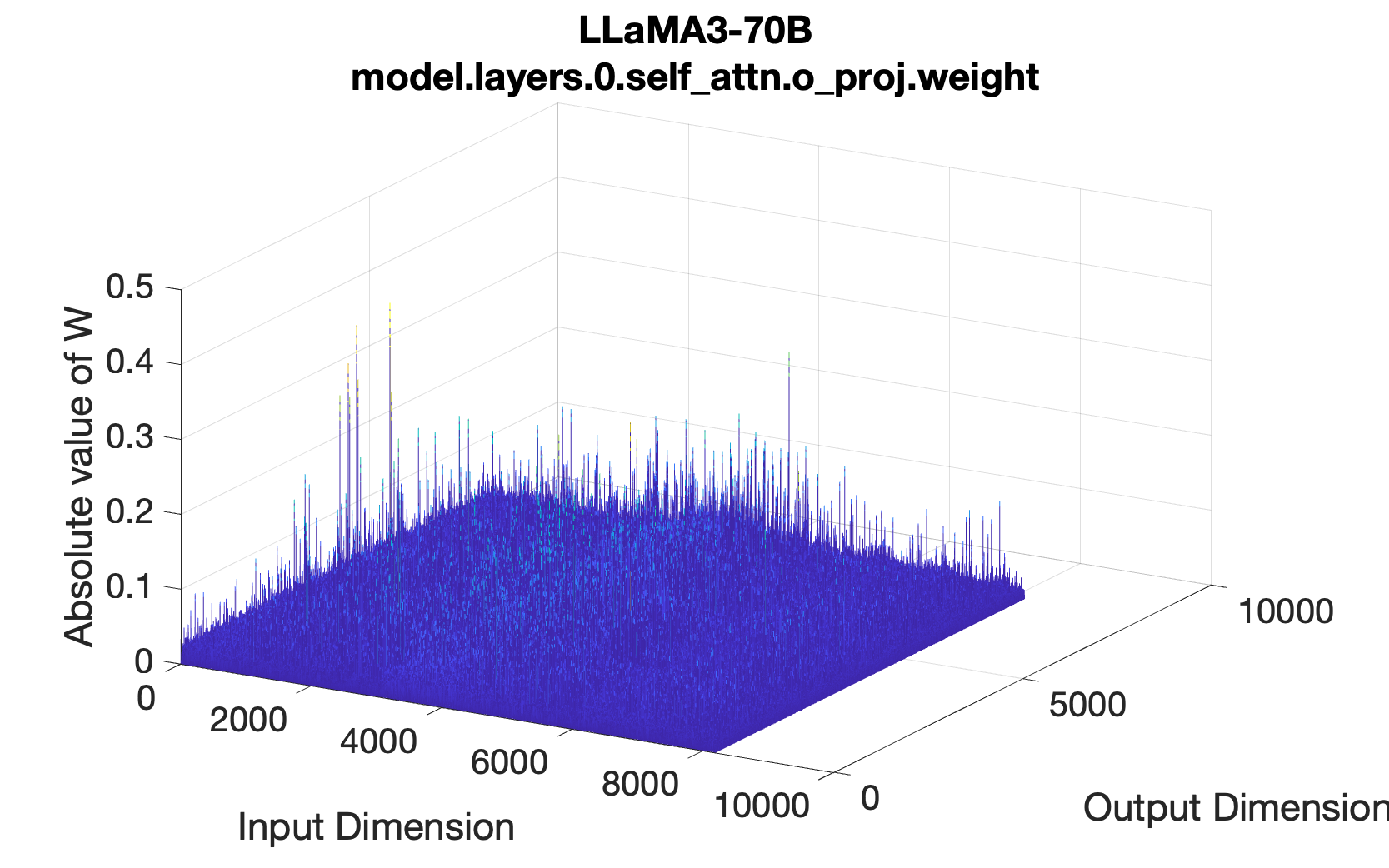}
    \caption{Weights in LLaMA3-70B }
    \label{fig:w_0_o}
\end{figure}

\begin{figure}[h]
    \centering
    \includegraphics[width=1\linewidth]{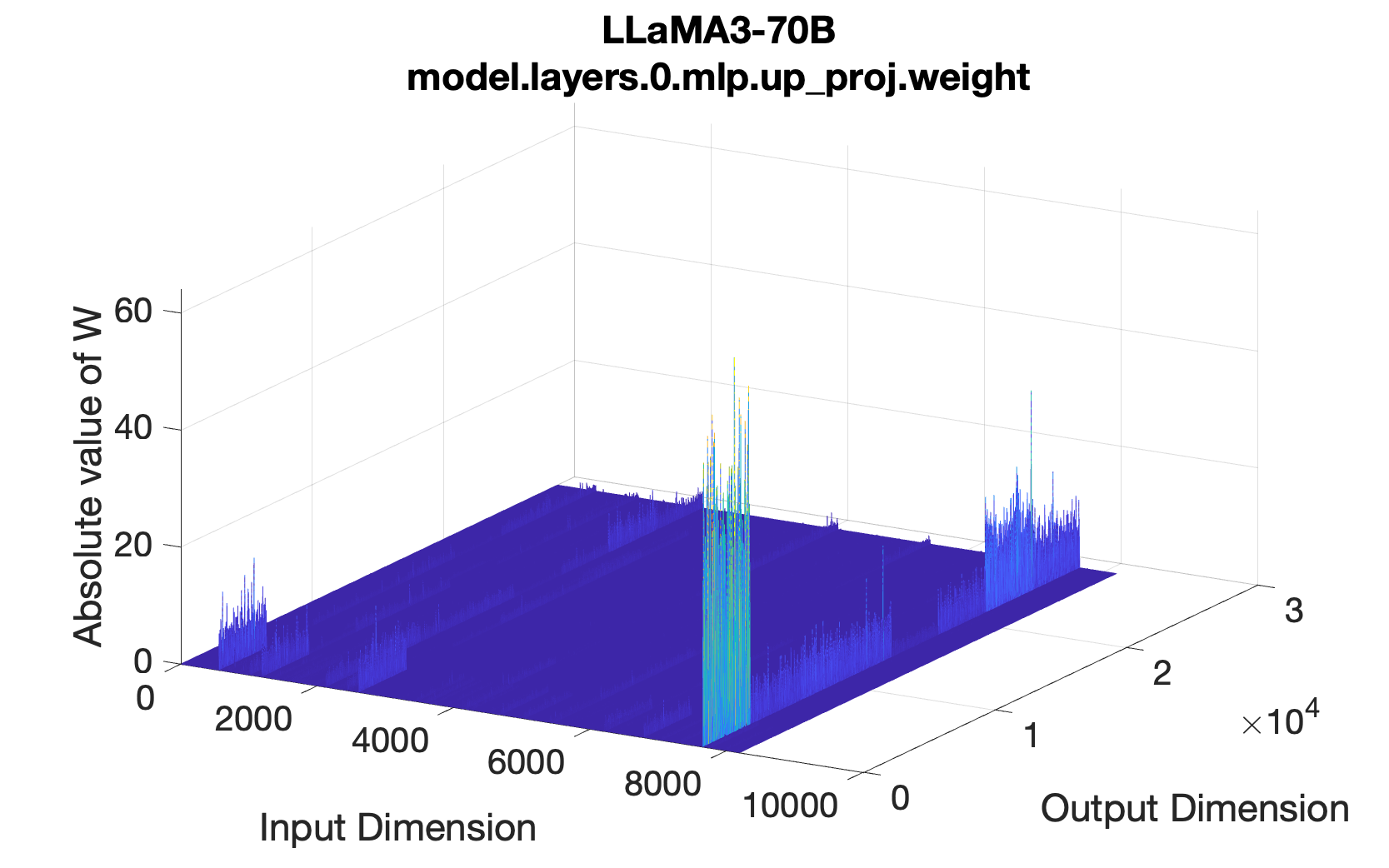}
    \caption{Weights in LLaMA3-70B }
    \label{fig:w_0_up}
\end{figure}

\begin{figure}[h]
    \centering
    \includegraphics[width=1\linewidth]{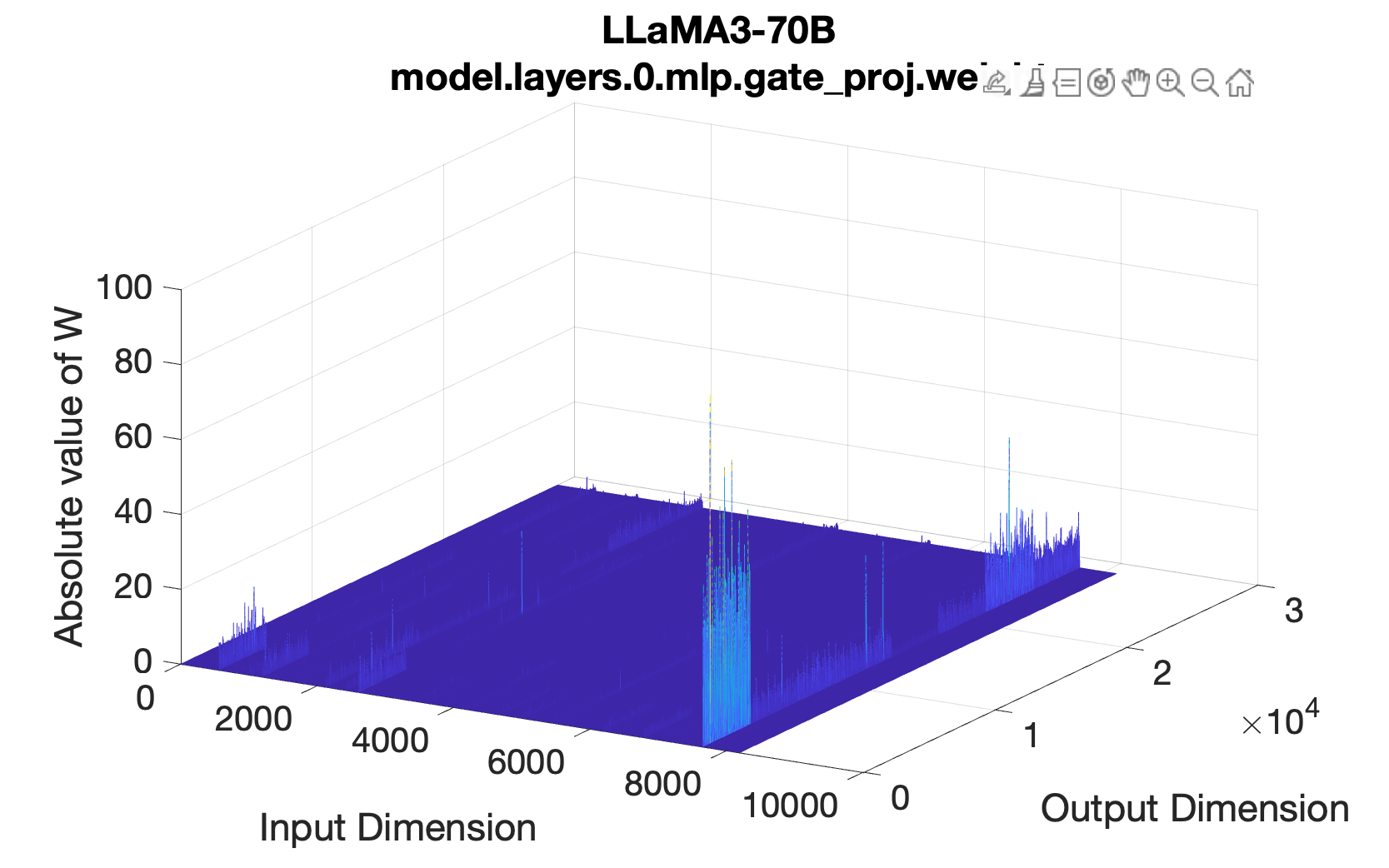}
    \caption{Weights in LLaMA3-70B }
\label{fig:w_0_gate}
\end{figure}

\begin{figure}[h]
    \centering
    \includegraphics[width=1\linewidth]{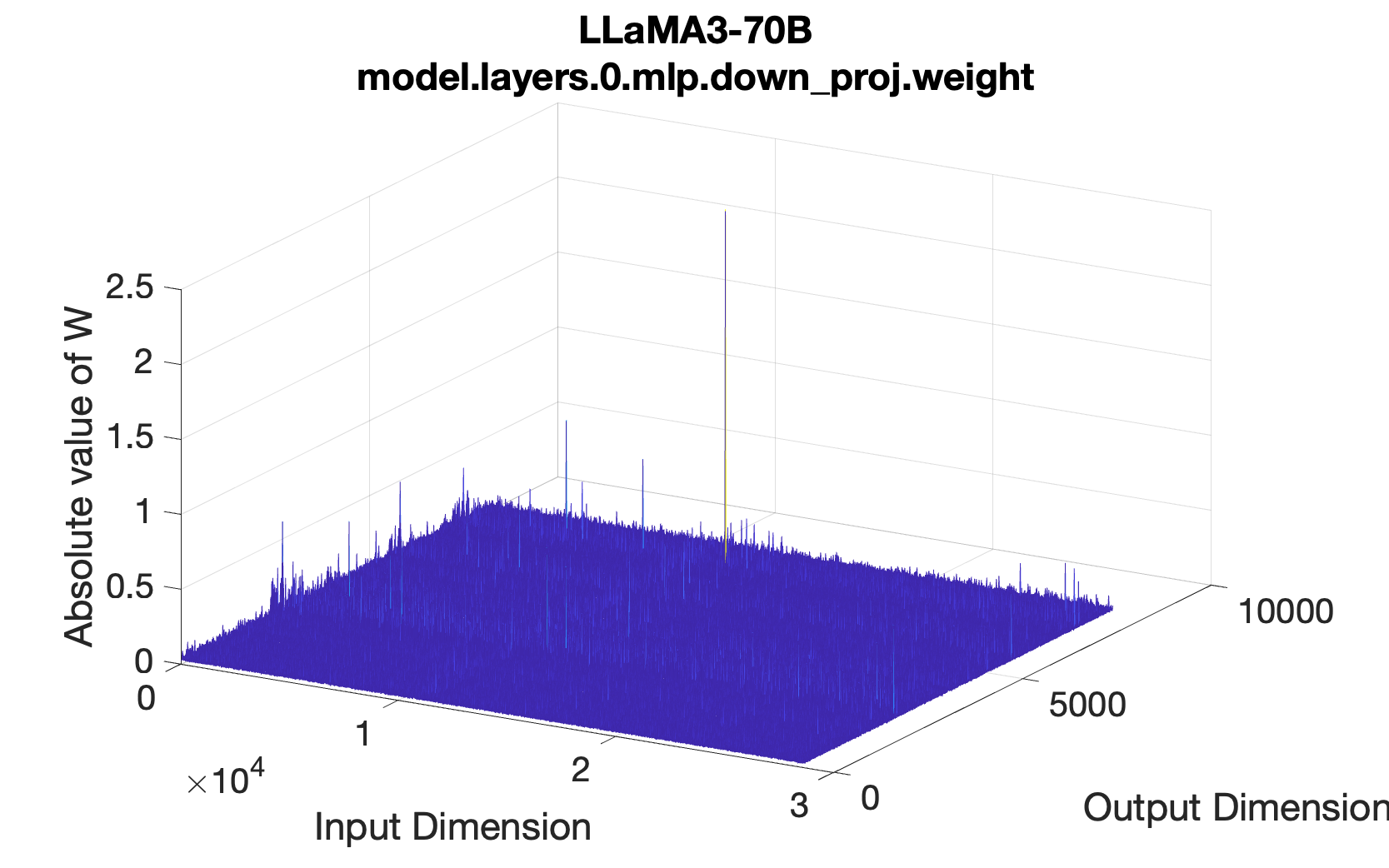}
    \caption{Weights in LLaMA3-70B }
    \label{fig:w_0_down}
\end{figure}

We show the distribution of weights in the first block in LLaMA3-70B. Figures~\ref{fig:w_0_q},~\ref{fig:w_0_k},~\ref{fig:w_0_v},~\ref{fig:w_0_o},~\ref{fig:w_0_up},~\ref{fig:w_0_gate},~\ref{fig:w_0_down} shows seven weight matrices in the 0-th Transformer block. The z-axis shows the magnitudes of outliers in these matrices. We can see that the Q, K, V, Up, and Gate matrices have larger outliers and they form a ``wall'' in some input indices. The other two weights, namely O and Down, do not show such a pattern. 


\subsubsection{The latter block has less large outliers}

\begin{figure}[h]
    \centering
    \includegraphics[width=1\linewidth]{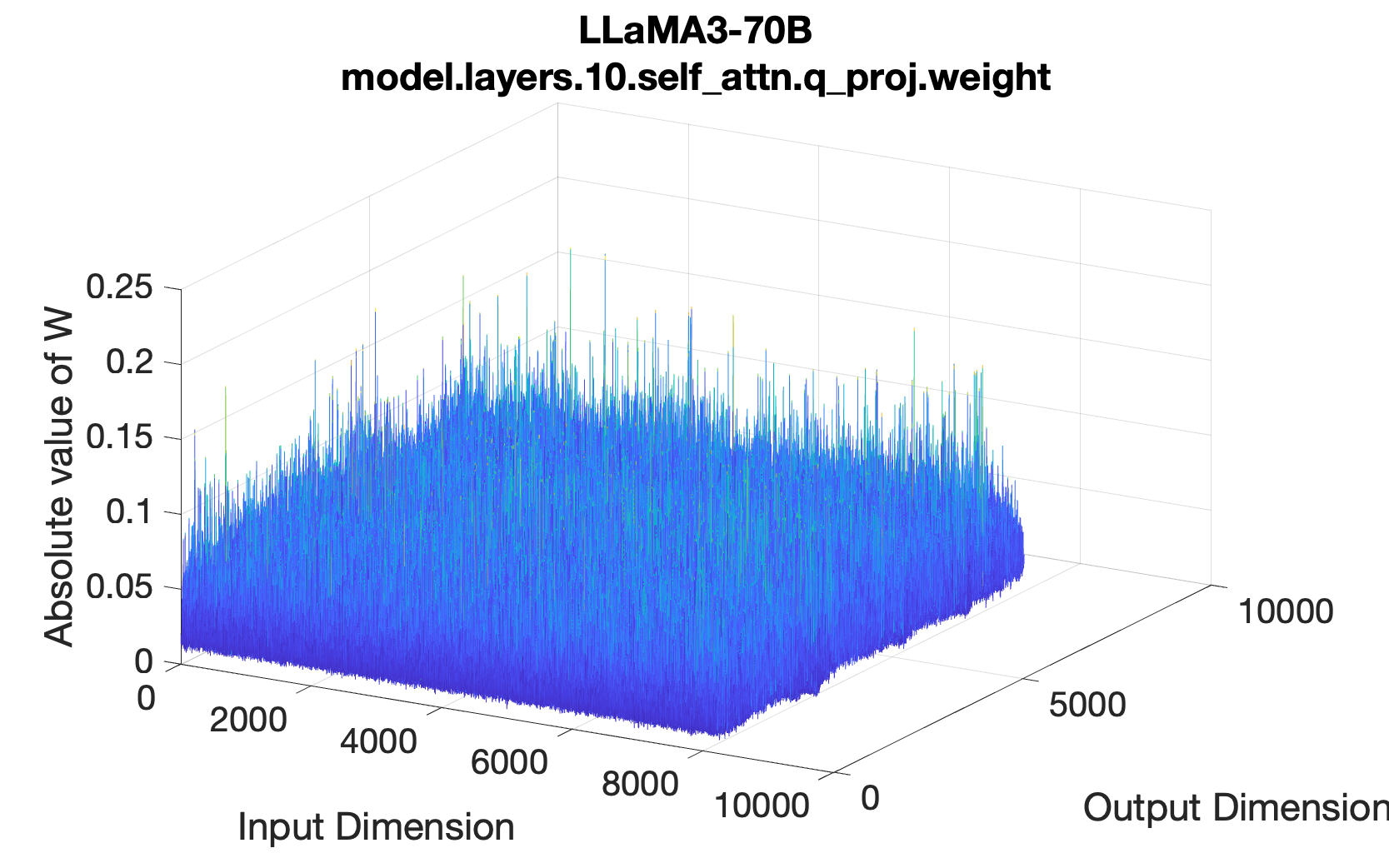}
    \caption{Weights in LLaMA3-70B }
    \label{fig:w_10_q}
\end{figure}

\begin{figure}[h]
    \centering
    \includegraphics[width=1\linewidth]{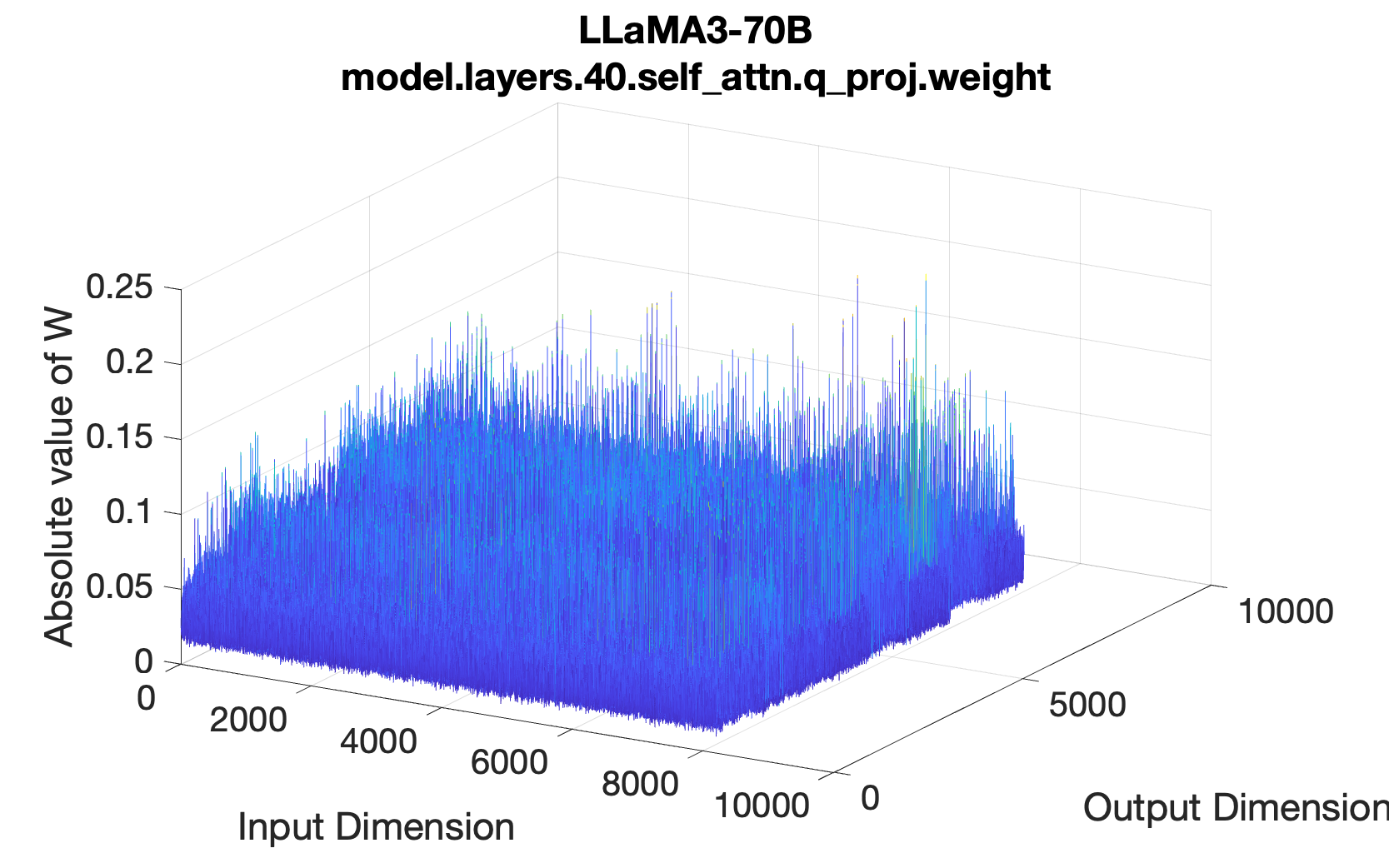}
    \caption{Weights in LLaMA3-70B }
    \label{fig:w_40_q}
\end{figure}

\begin{figure}[h]
    \centering
    \includegraphics[width=1\linewidth]{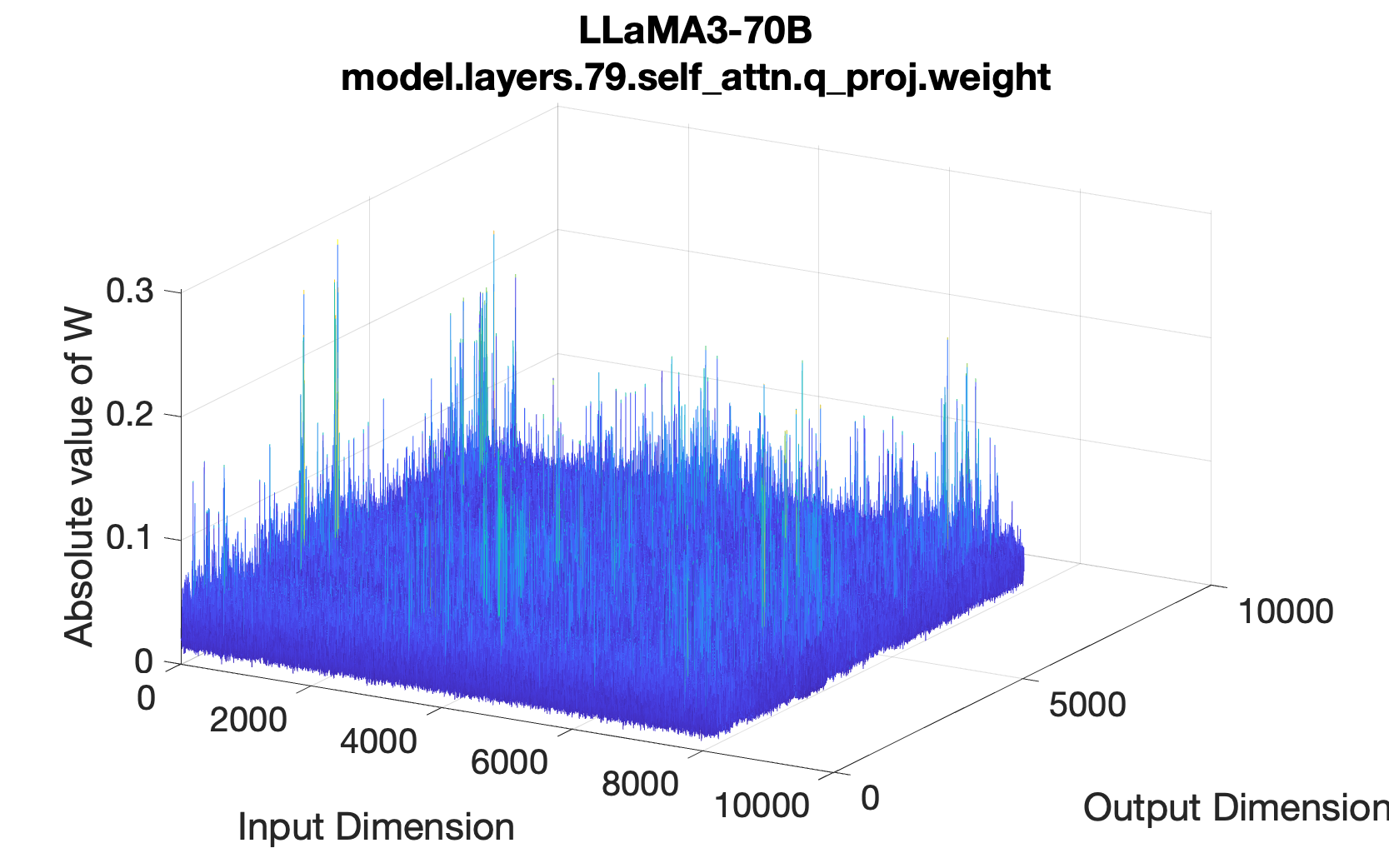}
    \caption{Weights in LLaMA3-70B }
    \label{fig:w_79_q}
\end{figure}

\begin{figure}[h]
    \centering
    \includegraphics[width=1\linewidth]{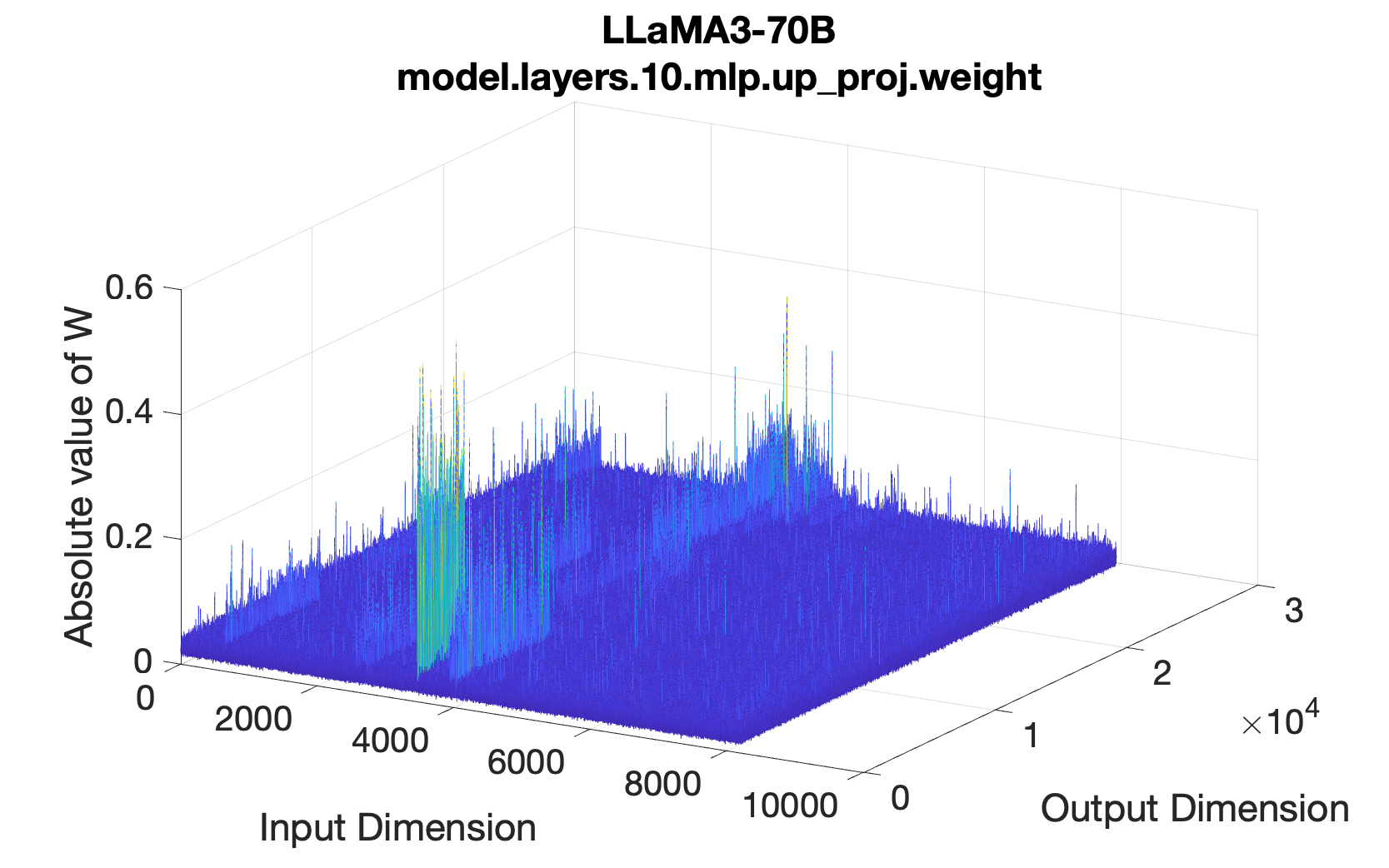}
    \caption{Weights in LLaMA3-70B }
    \label{fig:w_10_up}
\end{figure}

\begin{figure}[h]
    \centering
    \includegraphics[width=1\linewidth]{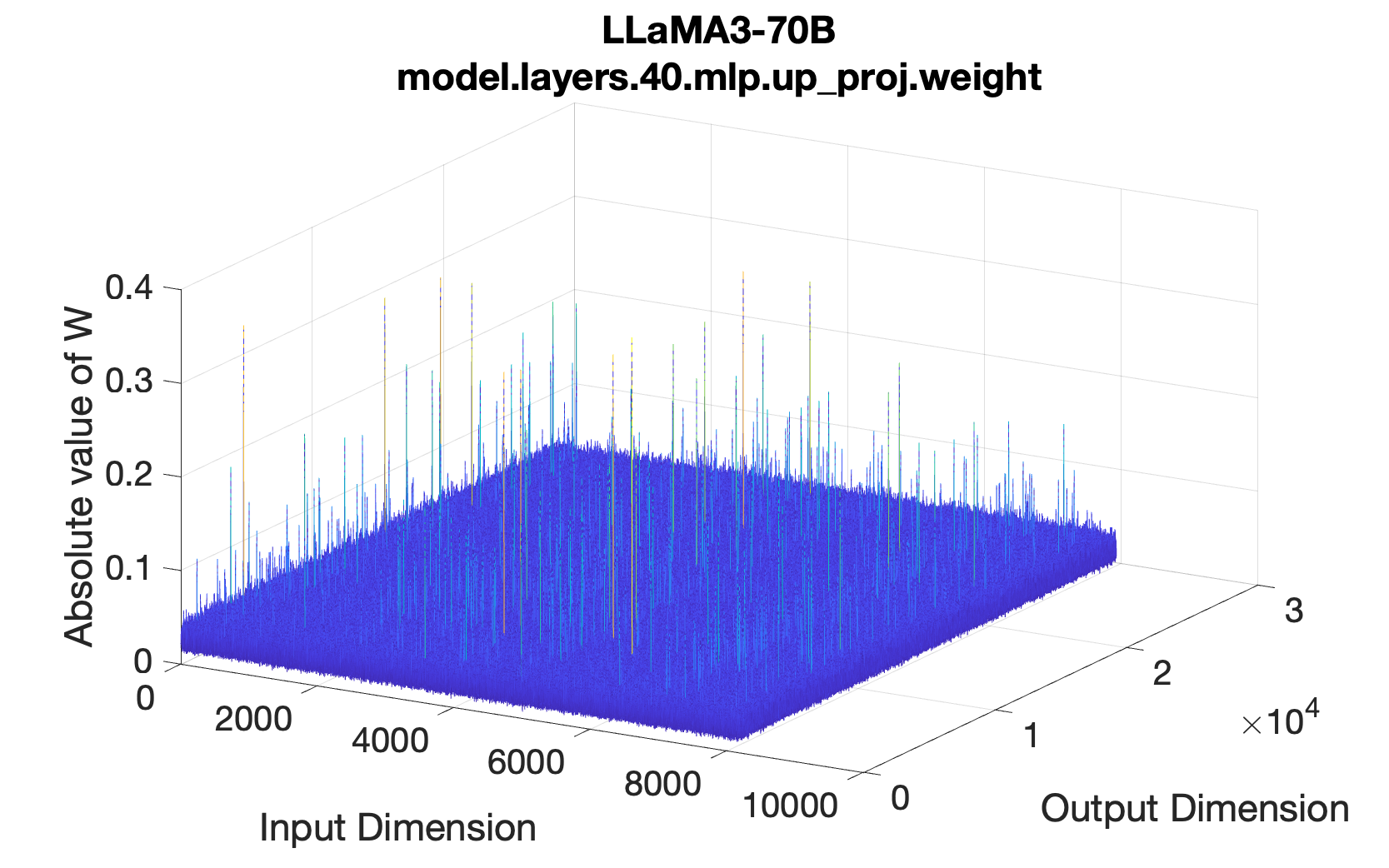}
    \caption{Weights in LLaMA3-70B }
    \label{fig:w_40_up}
\end{figure}

\begin{figure}[h]
    \centering
    \includegraphics[width=1\linewidth]{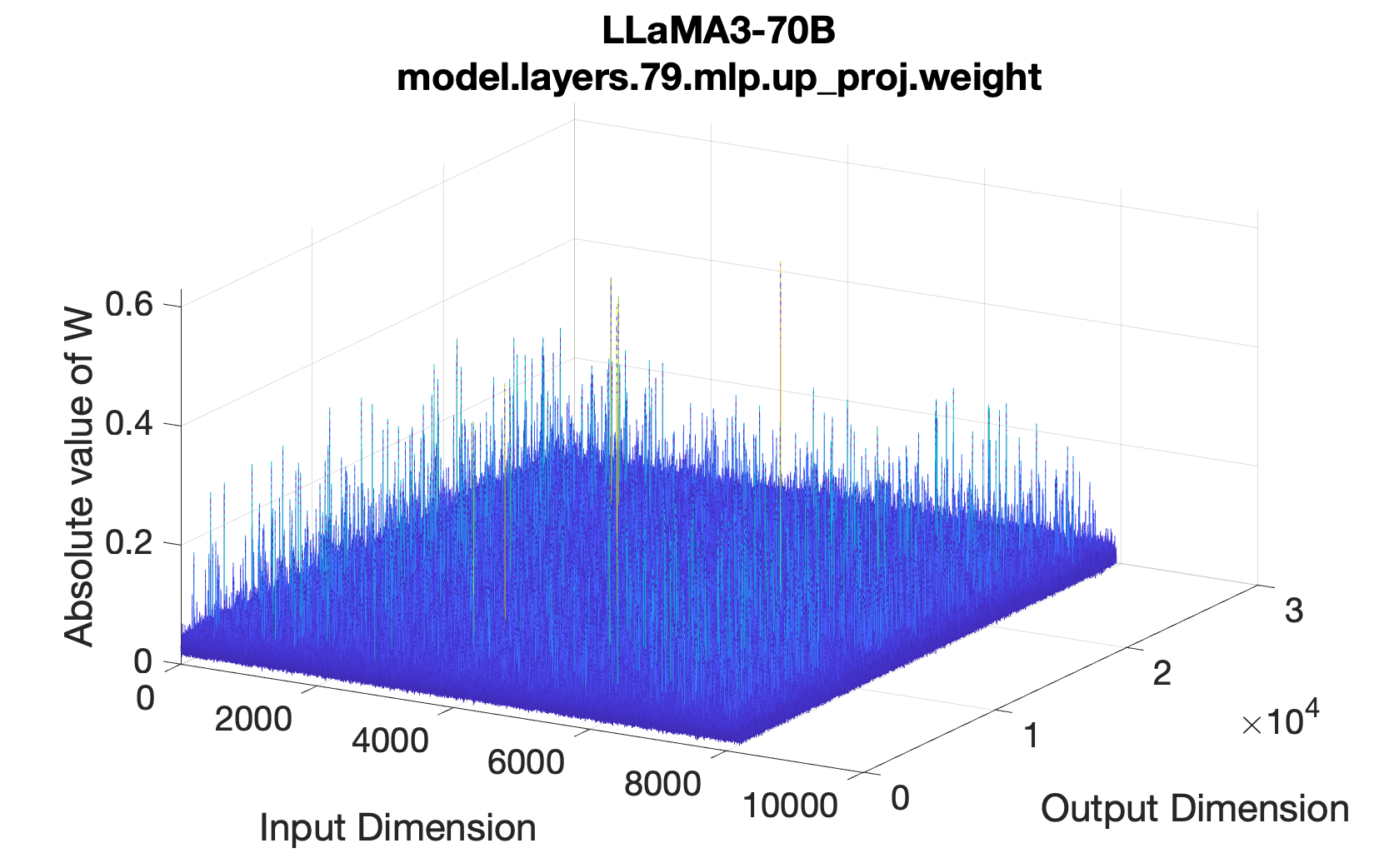}
    \caption{Weights in LLaMA3-70B }
    \label{fig:w_79_up}
\end{figure}

Figure~\ref{fig:w_10_q}, ~\ref{fig:w_40_q}, ~\ref{fig:w_79_q},~\ref{fig:w_10_up}, ~\ref{fig:w_40_up}, and~\ref{fig:w_79_up} show the Q and Up matrices in the 10th, 40th, and 79th (last) Transformer block in LLaMA3-70B. The magnitude of the weight outliers is significantly smaller than that in the first block, demonstrating more robustness to W8 per-channel quantization.


\subsection{Weights of Other LLM models}

\begin{figure}[h]
    \centering
    \includegraphics[width=1\linewidth]{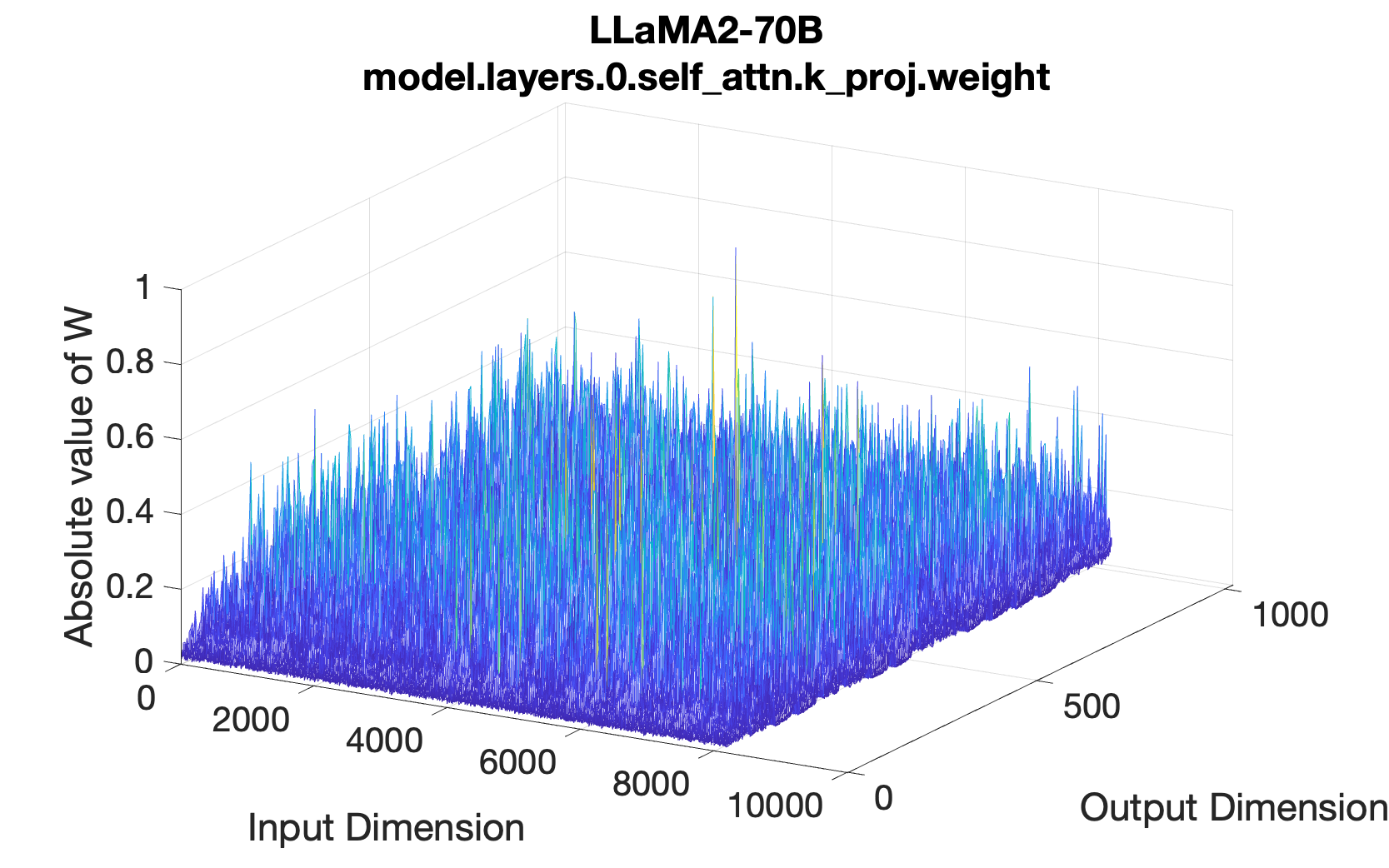}
    \caption{Weights with maximum quantization error of W8 in LLaMA2-70B, where max\_abs\_weight=0.95.}
    \label{fig:llama2-70b-L0-k}
\end{figure}

\begin{figure}[h]
    \centering
    \includegraphics[width=1\linewidth]{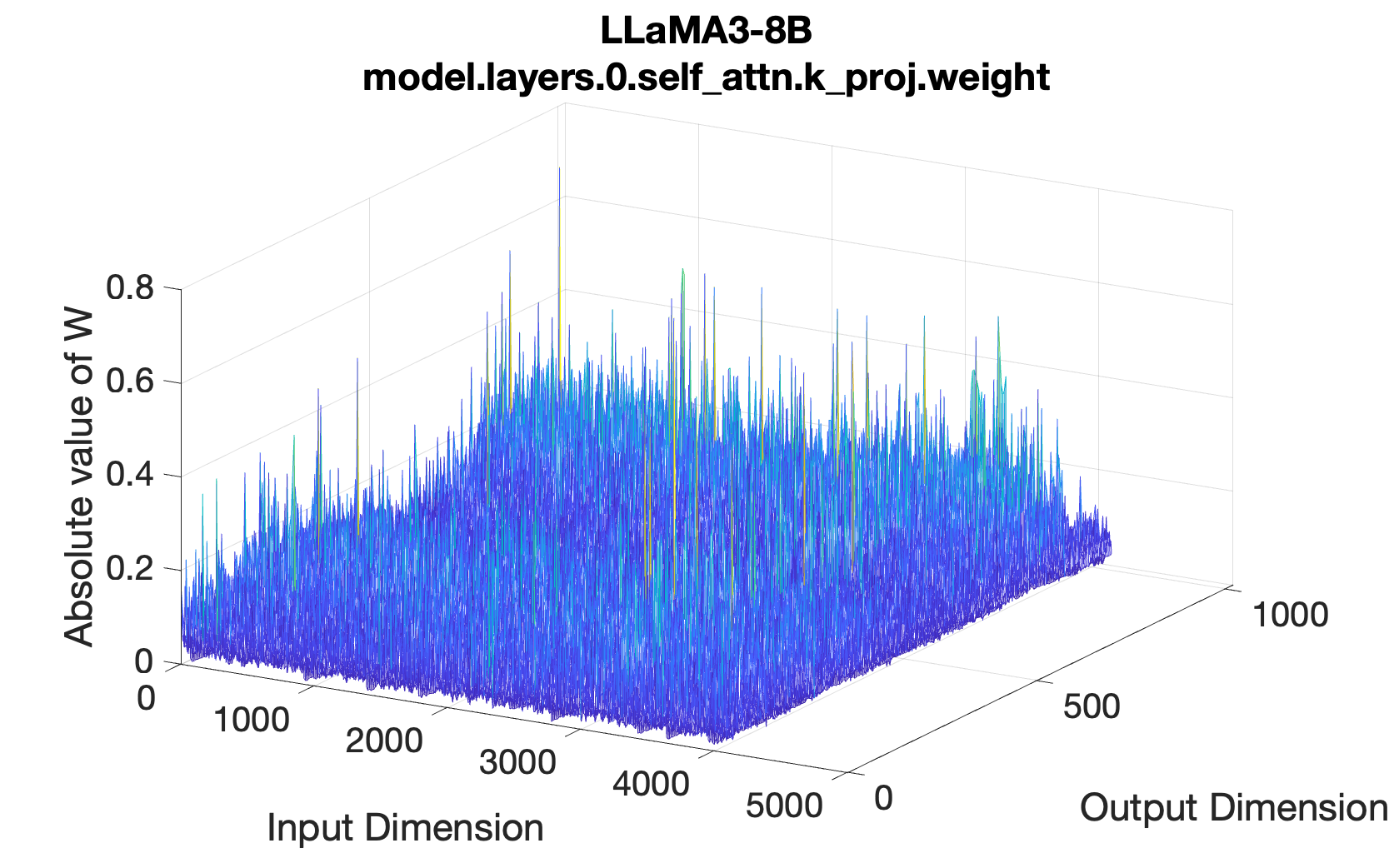}
    \caption{Weights with maximum quantization error of W8 in LLaMA3-8B, where max\_abs\_weight=0.77. }
    \label{fig:llama3-8b-L0-k}
\end{figure}

\begin{figure}[h]
    \centering
    \includegraphics[width=1\linewidth]{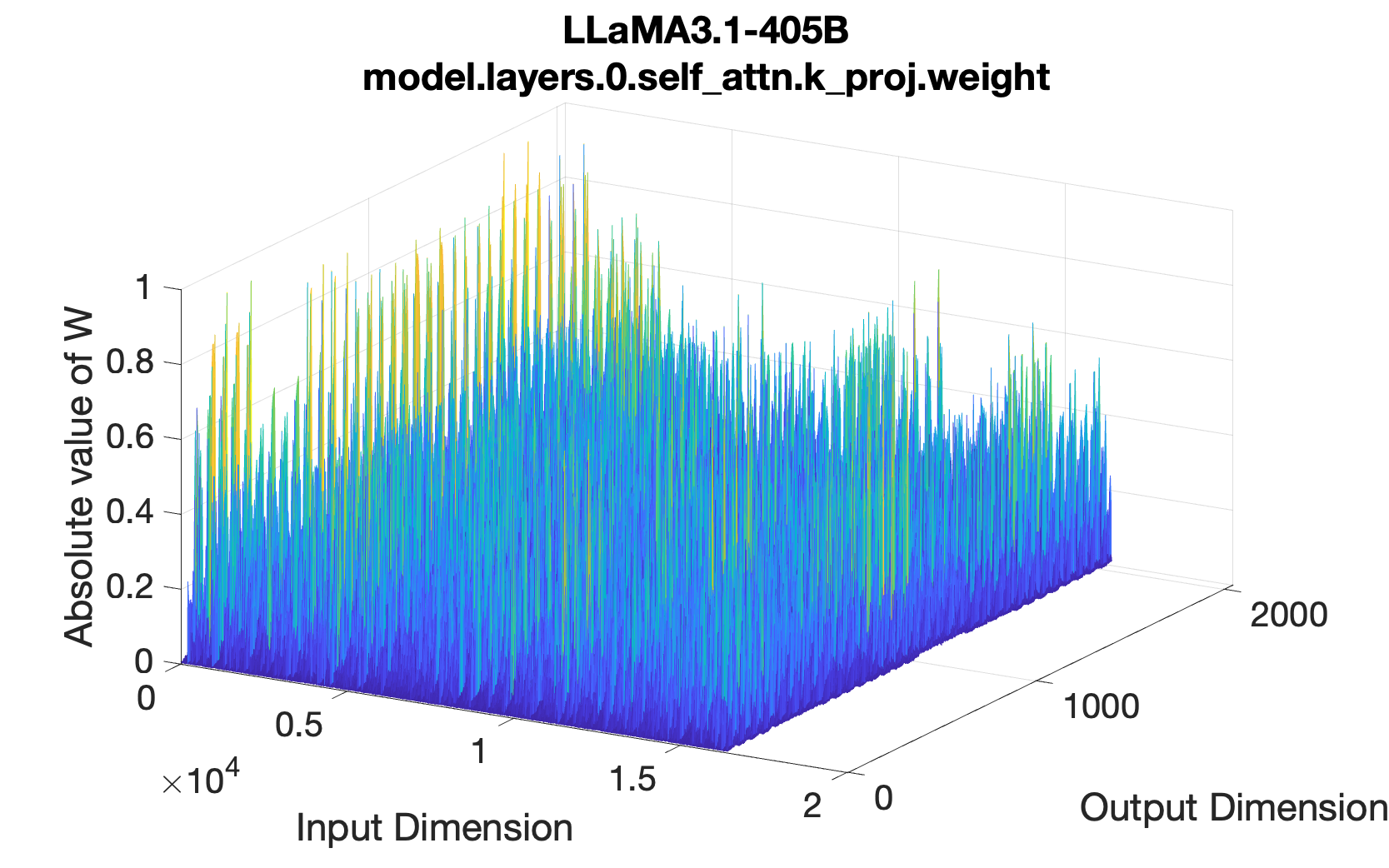}
    \caption{Weights with maximum quantization error of W8 in LLaMA3.1-405B, , where max\_abs\_weight=0.98. }
    \label{fig:llama3.1-405-L0-k}
\end{figure}

\begin{figure}[h]
    \centering
    \includegraphics[width=1\linewidth]{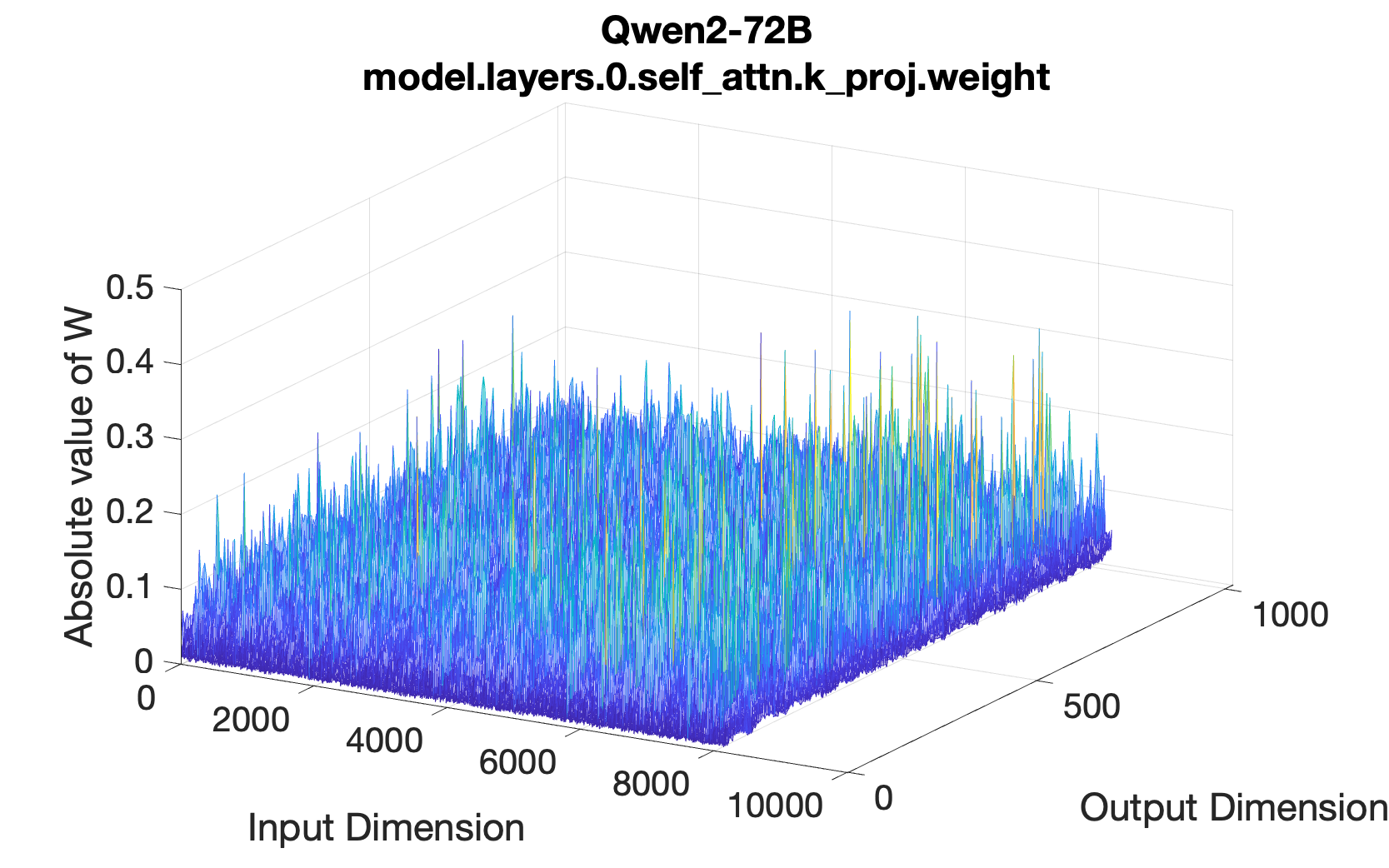}
    \caption{Weights with maximum quantization error of W8 in Qwen2-72B, , where max\_abs\_weight=0.41. }
    \label{fig:qwen2-72b-L0-k}
\end{figure}

Figure~\ref{fig:llama2-70b-L0-k}, ~\ref{fig:llama3-8b-L0-k}, ~\ref{fig:llama3.1-405-L0-k},~\ref{fig:qwen2-72b-L0-k} show the weights with maximum quantization error under W8 per-channel quantization in four LLMs, that is, LLaMA2-70B, LLaMA3-8B, LLaMA3.1-405B, and Qwen2-72B. It is worth noting that the maximum quantization error occurs all at the same layer, which is the ``K'' matrix in the first Transformer block. We can see that the weight outliers in the layer with maximum quantization error is less than 1.0, which is far smaller than that of LLaMA3-70B models, which is usually greater than 90. The difference of 2 orders of magnitude causes the quantization error of the LLaMA3-70Bs to be much more than the rest of the models.

\clearpage

\bibliographystyle{unsrtnat}

\end{document}